\documentclass[journal]{IEEEtran}

\usepackage{xcolor}
\usepackage{bm}
\usepackage{amsmath}
\usepackage{amsfonts}
\usepackage{amssymb}

\usepackage{amsthm}

\usepackage{mathtools}
\usepackage{tabularx,booktabs}
\usepackage{graphicx}
\usepackage{subfigure}
\usepackage{enumerate}
\usepackage{float}
\usepackage{url}

\usepackage{verbatim}
\usepackage{cite}
\usepackage{diagbox}
\usepackage{multirow}
\usepackage{makecell}
\usepackage{algorithm}  
\usepackage{algorithmicx}  
\usepackage{algpseudocode} 
\usepackage{pifont}
\usepackage[linkcolor=black,citecolor=black,urlcolor=blue,colorlinks=true]{hyperref}
\usepackage{stfloats}
\usepackage{booktabs} 
\usepackage{array}
\usepackage{pifont}

\graphicspath{{figures/}}
\IEEEoverridecommandlockouts

\definecolor{mygreen}{RGB}{0,130,0}
\definecolor{myred}{RGB}{230,0,0}

\author{ Longji Yin$^1$, Yunfan Ren$^1$, Fangcheng Zhu$^1$, Liuyu Shi$^1$, Fanze Kong$^1$,\\ Benxu Tang$^1$, Wenyi Liu$^1$, Ximin Lyu$^2$, and Fu Zhang$^1$
    \thanks{Corresponding author: Fu Zhang, {fuzhang@hku.hk.}}
    \thanks{$^1$The authors are with the Department of Mechanical Engineering, The University of Hong Kong, Hong Kong SAR, China. }
    \thanks{$^2$The author is with the School of Intelligent System Engineering, Sun Yat-sen University, Shenzhen, China.}
	
}

\title{Visibility-Aware Cooperative Tracking with Decentralized LiDAR-Based Aerial Swarms}
\begin{document}
    \maketitle

\begin{abstract}

Autonomous aerial tracking with drones offers vast potential for 
surveillance, cinematography, and industrial inspection. While 
single-drone tracking has been extensively studied, swarm-based 
target tracking remains underexplored despite its advantages in 
distributed perception, fault tolerance, and multidirectional 
coverage. We propose a decentralized LiDAR-based swarm tracking 
framework that enables visibility-aware cooperative tracking in 
complex environments. To address visibility, we introduce a novel Spherical Signed
Distance Field (SSDF)-based metric for 3-D environmental occlusion representation, coupled with an algorithm that
enables real-time onboard SSDF updating. A general Field-of-View (FOV) alignment cost supporting heterogeneous LiDAR configurations is proposed for consistent target observation. Swarm 
coordination is enhanced through costs that enforce inter-robot 
clearance, prevent mutual occlusions, and facilitate 3-D 
multidirectional target encirclement via a novel electrostatic-potential-inspired distribution metric. These are 
integrated into a hierarchical planner combining a kinodynamic 
front-end searcher with a spatiotemporal $SE(3)$ back-end 
optimizer. The proposed approach is thoroughly evaluated through comprehensive benchmark 
comparisons and ablation studies. Deployed on heterogeneous LiDAR 
swarms, the fully decentralized system supports collaborative 
perception, distributed planning, and dynamic reconfigurability. Validated through extensive real-world experiments in cluttered outdoor environments, the proposed system demonstrates robust cooperative tracking of dynamic targets (drones, humans) while achieving visibility maintenance. The code link is: \url{https://github.com/hku-mars/Swarm-Tracker}.

\end{abstract}


\section{Introduction}
\label{sec:intro}

\IEEEPARstart{A}{}utonomous aerial tracking with UAVs is now extensively applied in 
fields ranging from cinematography to surveillance and industrial 
inspection. While substantial progress has been made in single-UAV 
tracking, swarm-based aerial tracking remains underexplored. 
Cooperative UAV swarms surpass individual trackers by exploiting 
distributed perception, fault-tolerant redundancy, and 
multidirectional target coverage ~---~ features unattainable 
in single-UAV paradigms. In this work, we bridge this gap by 
introducing a new framework for visibility-aware target tracking 
with a cooperative team of decentralized UAVs.

\begin{figure}[tp] 
\vspace{+0.077cm}
\centering 
\includegraphics[width=1.0\linewidth]{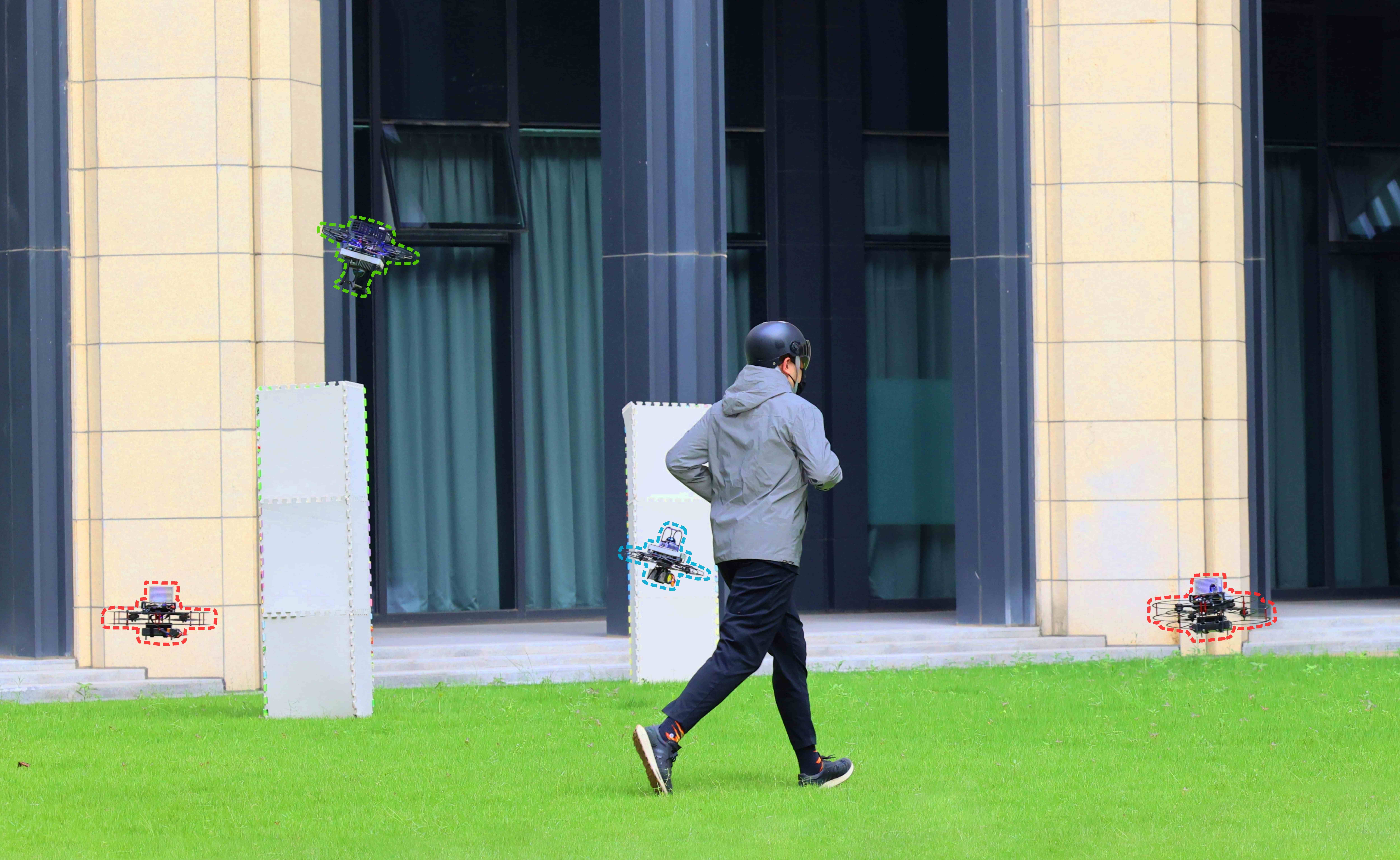} 
\caption{A swarm of four autonomous drones is cooperatively tracking a human runner using heterogeneous LiDAR configurations. The LiDAR setup consists of one upward-facing Mid360 LiDAR (marked by blue dashed lines), one downward-facing Mid360 LiDAR (green dashed lines), and two Avia LiDARs (red dashed lines). The swarm forms a 3-D distribution to track the target, with each tracker positioned optimally to suit its FOV settings. The video is at \url{https://www.youtube.com/watch?v=lTPE_JnsTPI}.} 
\label{fig: head_figure} 
\vspace{-0.45cm}
\end{figure}

Effective aerial tracking with autonomous swarms primarily relies on three criteria: visibility, coordination, and deployability. 
Visibility demands persistent sensor-based observation of the target 
during flight. Coordination requires the swarm to track cooperatively 
without impeding teammate performance. Deployability necessitates 
a complete and scalable decentralized system integration for real-world 
deployment. However, existing frameworks that effectively fulfill all three 
criteria are still lacking. Key technical challenges are outlined below.

The first challenge lies in designing accurate metrics to quantify 
visibility conditions in target tracking tasks. Line-of-sight (LOS) 
visibility is a well-established criterion in aerial tracking 
\cite{jeon2019online,jeon2020integrated, wang2021visibility, gao2023adaptive, lin2026eva,ji2022elastic, zhou2025control, ren2024intention}, requiring (1) an obstacle-free LOS between 
tracker and target, and (2) LOS alignment within the tracker's 
field-of-view (FOV). While various LOS-based visibility cost formulations 
have been proposed, existing solutions still exhibit notable 
deficiencies. The first LOS requisite needs a robust metric 
formulation to penalize occlusion and steer the LOS away from 
obstructed areas, but existing solutions are constrained by 
limitations such as non-differentiable 
formulations\cite{jeon2019online,jeon2020integrated}, inflexible 
dependencies on specific FOV 
shapes\cite{wang2021visibility, gao2023adaptive, lin2026eva}, and oversimplified 
2-D occlusion models\cite{ji2022elastic, zhou2025control, ren2024intention}. For the FOV requisite, 
many studies enforce fixed tracker-target altitude 
alignment\cite{zhou2022swarm, yin2023decentralized, lee2024dmvc}, 
a heuristic that disregards the 3-D FOV geometries and wastes the 
tracker's vertical agility. Current swarm tracking works \cite{zhou2022swarm, yin2023decentralized, lee2024dmvc, rao2025air, ho20213d, bucker2021you} rarely 
consider the heterogeneous FOV configurations' impact on swarm 
spatial distribution for target perception. Our work 
addresses these gaps through a novel SSDF-based visibility 
representation along with a unified FOV cost supporting heterogeneous 
sensor configurations.

The second challenge is coordinating the swarm's spatial 
distribution for effective 3-D target tracking. Existing multi-UAV 
tracking frameworks \cite{zhou2022swarm, yin2023decentralized, lee2024dmvc, liu2023formation2, liu2025formation, tallamraju2019active} mainly adopt simplistic 2-D equidistant leader-follower formations that severely waste the swarm's 3-D maneuverability. However, 
the swarm needs to optimize its 3-D spatial distribution to 
achieve multidirectional target coverage, which can mitigate 
single-direction occlusions, support target measurement fusion 
from diverse perspectives, and grant each tracker a larger angular 
space to respond to adverse situations. This multidirectional 
approach can also maximize the utility of heterogeneous sensors 
by positioning agents at vantage points suited to their FOV 
modalities. We address this gap by proposing a novel 3-D 
multidirectional tracker distribution formulation. In addition, 
swarm coordination also requires preventing mutual LOS occlusion 
among teammates and ensuring inter-agent safety during cooperative 
tracking.

The third challenge is the system-level integration of a 
decentralized swarm tracking system. A multi-UAV tracking system should be capable of coordinated 
planning under a decentralized architecture, avoiding single 
points of failure inherent in centralized designs. 
Moreover, capabilities including shared estimation of passive dynamic targets and online map synchronization are essential for practical applicability in unknown complex outdoor environments. Support for heterogeneous FOV configurations is also an important consideration for multi-UAV 
tracking systems. However, existing work \cite{zhou2022swarm, yin2023decentralized, lee2024dmvc, 
tallamraju2019active, liu2023formation2, liu2025formation, rao2025air, ho20213d, bucker2021you} that demonstrates all these 
integrated capabilities in real-world outdoor scenarios is still 
lacking.

Based on the above analysis, we propose a complete swarm tracking system that addresses all the outlined challenges. To resolve the LOS visibility challenge, we novelly adapt the SSDF, 
a spatial representation originally used for graphical 
rendering\cite{wang2009all, wang2012analytic,wang2014parallel,
iwasaki2014interactive}, into a differentiable visibility model 
tailored for aerial tracking, encoding the 3-D environmental 
occlusion around the target. To the best of our knowledge, this is 
the first application of SSDF in robotic planning. 
To support high-frequency onboard computation, we devise an SSDF update algorithm exploiting the monotonic property 
of LOS visibility. To ensure FOV alignments, we formulate a unified FOV constraint supporting heterogeneous FOV settings. To enable 3-D multidirectional target encirclement, we draw 
inspiration from Thomson's classical electron distribution problem, 
novelly framing the optimal swarm distribution as electrostatic potential 
minimization for the first time in multi-UAV tracking. Additional metrics for the tracking distance, obstacle avoidance, dynamic feasibility, mutual occlusion avoidance, and inter-agent collision avoidance are also formulated to assess the tracking motion.

To translate the proposed metrics into tracking motions, 
we present a hierarchical trajectory generation framework consisting 
of a front-end kinodynamic searcher and a back-end spatiotemporal 
optimizer. Both stages incorporate the metrics into their respective 
formulations. To handle the 
newly introduced higher-order constraints such as FOV alignment, 
the back-end performs $SE(3)$ full-state optimization. Yaw trajectory is also jointly 
formulated into the back-end to support varied FOV 
configurations. At the system level, we present a fully decentralized swarm 
architecture integrating localization, mapping, planning, and 
control modules. Trajectory synchronization, collaborative target estimation, and 
map sharing are implemented across the swarm to enable tight 
coordination and collective perception. We validate the system through comparative simulation 
benchmarks and extensive real-world experiments in complex 
unknown outdoor environments, demonstrating visibility-aware 
cooperative tracking of passive dynamic targets (drones, human 
runners) with heterogeneous LiDAR FOVs and dynamic swarm reconfigurability.

The contributions of this paper are summarized as:
\begin{enumerate}

 \item An SSDF-based 3-D occlusion representation is 
    introduced for aerial tracking, with a proposed 3-D SSDF 
    update algorithm for real-time onboard SSDF computation.

\item Differentiable visibility metrics are developed, including an SSDF-based occlusion penalty and a general FOV alignment cost supporting heterogeneous FOV settings.

\item A 3-D swarm distribution metric is formulated for 
cooperative aerial tracking, based on the electrostatic 
Thomson's problem and enabling 3-D multidirectional target 
encirclement.
    
\item A complete decentralized multi-UAV tracking system is 
presented, integrating a two-stage planning framework (a 
kinodynamic front-end searcher and an $SE(3)$ back-end 
optimizer), shared target estimation, map synchronization, 
and dynamic reconfigurability. The system is validated through 
extensive real-world experiments in complex outdoor 
environments, and the source code is publicly released.
\end{enumerate}

In what follows, Sec.~\ref{sec:related works} 
reviews related works and Sec.~\ref{sec:system overview and preliminaries} provides the system 
overview and preliminaries. Sec.~\ref{sec:spherical_signed_distance_field} introduces the 
SSDF-based visibility model and update algorithms. Planning metrics 
are formulated in Sec.~\ref{sec:visibility_aware_swarm_tracking}, followed 
by front-end and back-end introduction in Secs.~\ref{sec:kinodynamic_searching} 
and \ref{sec:spatial_temporal_trajectory_optimization}. Benchmark and experimental 
results are presented in Secs.~\ref{sec:benchmark} and \ref{sec:experiment}. 
Sec.~\ref{sec:conclusion} provides discussion and conclusion.

\section{Related works}
\label{sec:related works}
\subsection{Single-UAV Target Tracking}
\label{Single-UAV target tracking}

Several earlier studies \cite{kendall2014board, cheng2017autonomous} treat 
vision-based aerial tracking as a local control problem, but struggle to 
account for visibility constraints or plan for future target motions. 
Han \textit{et al.} \cite{han2021fast} proposed an optimization-based 
tracking planner consisting of a spatiotemporal optimizer and kinodynamic 
searcher, but they only focus on maintaining tracking distance while 
neglecting visibility. Penin \textit{et al.} \cite{penin2018vision} 
formulate a nonlinear MPC penalizing occlusion, but assume ellipsoid-shaped 
obstacles. Bonatti \textit{et al.} \cite{bonatti2020autonomous} developed 
an aerial cinematography framework considering LOS visibility, but ignored 
the sensor's FOV geometry. Wang \textit{et al.} \cite{wang2021visibility} 
design an occlusion cost that penalizes obstacle-FOV intersections. However, the proposed constraint is too strict to be satisfied in dense spaces. Besides, their cost is specifically formulated for conic FOVs and not compatible with omnidirectional sensors like 360$^\circ$
LiDARs, as the obstacle-intersection area can always exist in the omnidirectional view. Jeon \textit{et al.} \cite{jeon2020integrated} present an Euclidean Signed Distance Field (ESDF) based 
visibility metric, assessing the occlusion by evaluating the minimum ESDF value along the LOS, but it lacks differentiability for back-end optimization. Zhou \textit{et al.}\cite{zhou2025control} model visibility-awareness as a control barrier function (CBF) constraint using the 
signed distance to the occluded FOV, enabling reactive 
pursuit-evasion control under occlusion.
Ji \textit{et al.} \cite{ji2022elastic} generate 2-D sector-shaped visible 
regions by ray-casting, but these 2-D sectors represent visibility only at a specific height, which is unable to facilitate 3-D occlusion avoidance. 
Additionally, the seed for building visibility sectors must be in visible 
areas, unable to guide trackers when already occluded. To systematically address 
these limitations, we adopt SSDFs \cite{michikawa2008spherical, wang2009all, 
wang2012analytic, wang2014parallel} to encode 3-D spatial visibility around 
the target. Our formulation is differentiable, independent of FOV type, and 
models full 3-D visibility without seed restrictions.

\subsection{Multiple-UAV Target Tracking}
\label{Multi-UAV target tracking}

Motion planning for multi-UAV target tracking has gained growing 
interest \cite{zhou2022swarm, yin2023decentralized,tallamraju2019active, bucker2021you, 
ho20213d, lee2024dmvc, nageli2017real2, rao2025air}. 
Zhou \textit{et al.}\cite{zhou2022swarm} present a swarm using a 
fixed 2-D leader-follower formation, sharing target position to 
enhance occlusion resistance. Tallamraju \textit{et al.}\cite{tallamraju2019active} 
employ a 2-D equidistant formation with MPC for formation control 
and obstacle avoidance. While \cite{zhou2022swarm, tallamraju2019active} 
share target observations to endure occlusion, they lack strategies 
to proactively mitigate visibility loss. Nageli \textit{et al.}\cite{nageli2017real2} 
propose an MPC-based framework using horizon planes to separate 
visible and invisible regions, but assume ellipsoidal obstacles. 
Lee \textit{et al.}\cite{lee2024dmvc} employ 2-D Inter-Visibility 
Cells to prevent occlusion, but only consider visibility at the 
front-end with pure path smoothing in the back-end, leading to 
potential constraint violations. Bucker \textit{et al.}\cite{bucker2021you} 
use discrete cells with occlusion scores and a centralized greedy 
algorithm, but the predefined priority can result in sub-optimal 
outcomes. Ho \textit{et al.}\cite{ho20213d} propose a 2-D formation 
rotating around the target using centralized dynamic programming, 
but the fixed formation is prone to failure in cluttered environments 
and susceptible to node failures.

Multi-robot multi-target tracking (MR-MTT) has received 
increasing attention. Schlotfeldt 
\textit{et al.}~\cite{schlotfeldt2018anytime} proposed anytime 
decentralized planning with distributed information gathering. 
Ramachandran \textit{et al.}~\cite{ramachandran2023resilient} 
addressed visual sensor degradation via communication topology 
reconfiguration. Li 
\textit{et al.}~\cite{li2025resilient, li2025failure} proposed 
chance-constrained optimization for sensing and communication 
failures. Yu \textit{et al.}~\cite{yu2025dronefl} developed 
federated learning with altitude-based normalization. While 
these MR-MTT works focus on coverage and estimation accuracy 
across multiple targets, our work targets the persistent 
visibility of a single target through occlusion-aware 
cooperative planning.

More recently, Yin \textit{et al.}~\cite{yin2023decentralized} 
proposed a decentralized swarm tracking framework. However, 
several limitations remain. First, they adopt the 2-D 
sector-based visibility metric from Ji 
\textit{et al.}~\cite{ji2022elastic}, which cannot handle 3-D 
occlusion; our SSDF-based method addresses this limitation. 
Second, their framework lacks explicit FOV modeling and rigidly 
aligns the tracker and target at the same altitude, making it 
incompatible with FOV-limited LiDARs; our framework models 
generic 3-D FOV constraints. Moreover, similar to other 
multi-UAV tracking works~\cite{zhou2022swarm, lee2024dmvc, 
liu2023formation2, liu2025formation}, Yin \textit{et al.} 
suggest a 2-D equidistant formation to encircle the target. 
However, such formations underutilize the 3-D maneuverability 
of UAVs and may conflict with other tracking requirements. To 
address this, we propose an electrostatic-potential-inspired 
3-D distribution metric for the coordination. For the 
optimization, \cite{yin2023decentralized} lacks $SE(3)$ ability 
and terminal state optimization, and does not formulate yaw 
trajectory into the back-end. At the integration level, the 
trackers in~\cite{yin2023decentralized} rely on target 
communication for target state acquisition, making it 
applicable only to communicative targets. Our system 
demonstrates tracking of non-cooperative passive targets with a 
more complete decentralized architecture with team data sharing.

\begin{figure*}[!t]
  \centering
  \includegraphics[width=1.0\textwidth]{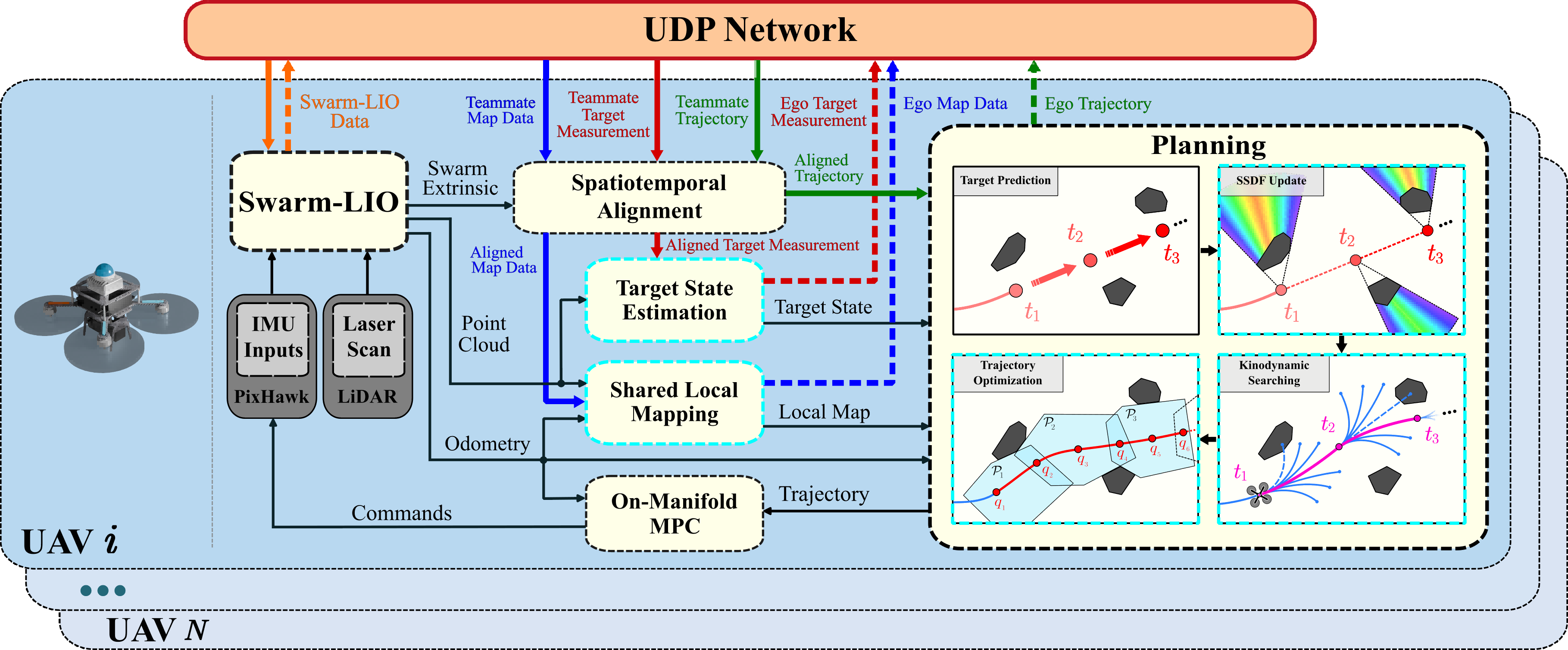}
  \caption{An overview of our decentralized swarm tracking system, 
including decentralized swarm localization, shared mapping, 
collaborative target estimation, onboard control, and motion 
planning modules. A hierarchical planner is designed to 
generate optimal trajectories for swarm tracking. All critical 
data is exchanged between swarm members via a UDP-based 
wireless network. The newly added or improved blocks are marked by cyan dashed curves.}
  \label{fig:system_architecture}
  \vspace{-0.2cm}
\end{figure*}

\section{System Overview and Preliminaries}
\label{sec:system overview and preliminaries}

\subsection{System Overview}
\label{sec:system overview}
We begin by stating the cooperative target tracking problem. Consider a 
decentralized swarm of $N$ UAVs tracking a passive target in an unknown 
cluttered environment. Each UAV $i$ is equipped with a LiDAR sensor, 
and the swarm communicates via a wireless network to exchange trajectories, 
map updates, and target observations. Each UAV $i$ has access to its 
ego-state estimate $\mathbf{x}_i$ from onboard LiDAR-inertial odometry, 
a local occupancy map $\mathcal{M}_i$, 
fused target states $\mathbf{x}_{\xi}$, and teammates' broadcast trajectories 
$\{\mathbf{p}_j(t)\}_{j \neq i}$. The objective is to compute, for each 
UAV $i$, a trajectory $\mathbf{p}_i(t): [0, T_p] \rightarrow \mathbb{R}^3$ 
that solves:
\begin{equation}
    \min_{\mathbf{p}_i(t)} \; \mathcal{J}(\mathbf{p}_i,  \mathbf{x}_{\xi}, \mathcal{M}_i, \{\mathbf{p}_j\}_{j \neq i}),
\end{equation}
where $\mathcal{J}$ encapsulates the visibility-aware tracking objectives 
and swarm coordination costs discussed in Sec.~I, subject to safety and dynamic feasibility 
constraints. The formulation of $\mathcal{J}$ and the planning framework 
to solve this problem are the main contributions, detailed in Secs.~IV--VII.

Underpinning this planning solution is a decentralized tightly-integrated 
swarm system that provides each UAV with the necessary information, as illustrated in Fig.~\ref{fig:system_architecture}. Each 
drone runs Swarm-LIO \cite{zhu2024swarm} for self-localization and 
mutual state estimation, with all shared data spatiotemporally aligned 
using calibrated swarm extrinsics and time offsets. Three types of 
information are exchanged among teammates: (1) planned trajectories for 
coordination, (2) local map updates \cite{ren2024rog, shi2024real} for 
collective environmental perception, and (3) target measurements for 
multi-source state estimation. More implementation details are presented in 
Sec.~\ref{sec:System Setup and Implementation Detail}.

Given the information from the system, we solve the trajectory generation in a 
hierarchical manner. The prediction 
module first extrapolates future target positions from the fused target states.
An SSDF is then built at each predicted target position to model 3-D 
occlusion and enable visibility-awareness (Sec.~\ref{sec:spherical_signed_distance_field}). 
With the tracking objectives in $\mathcal{J}
$ collectively formulated, 
we employ a kinodynamic searcher to expand motion primitives and select 
an optimal path (Sec.~\ref{sec:kinodynamic_searching}), used for safe 
corridor generation and back-end trajectory initialization. The trajectory's 
spatial and temporal profiles are then optimized by the back-end to 
maximize tracking performance (Sec.~\ref{sec:spatial_temporal_trajectory_optimization}). 
This planning process cycles periodically with a receding horizon, and 
the generated trajectories are executed via a model predictive controller 
\cite{lu2022manifold}.

\subsection{Target State Estimation}
\label{sec:target state estimation}

The target is detected using LiDAR point clouds. Following \cite{zhu2024swarm}, we employ Euclidean clustering to 
process the target points, and the centroid of the clustered points 
serves as the target position measurement. In our experiments, 
high-reflectivity markers are attached to the target, allowing us 
to filter out target points based on reflectivity information.

In this work, the target is modeled as a kinematic point. The target state 
vector $\mathbf{x} = [\mathbf{p}^T, \mathbf{v}^T]^T \in \mathbb{R}^6$, 
consisting of position and velocity, is estimated using a 
constant-velocity error-state Kalman filter (ESKF). The motion model is
\begin{equation}
    \mathbf{x}_{k+1} = \mathbf{F} \mathbf{x}_k + \mathbf{w}_k, \quad
    \mathbf{F} = \begin{bmatrix} \mathbf{I}_3 & \Delta t \cdot \mathbf{I}_3 \\ 
    \mathbf{0}_3 & \mathbf{I}_3 \end{bmatrix},
\end{equation}
\noindent where $\mathbf{w}_k$ is the process noise. Each drone runs its own ESKF and detects the target position independently. The measurements are shared among 
teammates, and each drone sequentially fuses the measurements (ego and received) using the standard ESKF update equations. This decentralized fusion scheme ensures that each drone benefits from multi-viewpoint observations while maintaining autonomy. The unimodal 
Kalman filter cannot represent multiple hypotheses during 
prolonged occlusion; however, our system actively prevents such 
scenarios through visibility-aware planning and decentralized 
fusion, making the ESKF sufficient in practice.

\subsection{Target Motion Prediction}
\label{sec:target motion prediction}
The prediction module takes the target state from the ESKF as input and 
generates a sequence of future positions $\{\bm{\xi}_k\}$ via 
constant-velocity linear extrapolation: $\bm{\xi}_k = \mathbf{p} 
+ \mathbf{v} \cdot k \cdot \delta T$, for $0 \leq k \leq N_p$, 
where $\delta T$ is the time step and $N_p$ is the number of prediction steps. 
The predicted sequence is denoted as
\begin{equation}\label{eqn:q_target_set}
    Q_{target} = \{\bm{\xi}_k 
    \in \mathbb{R}^3 \mid 0 \leq k \leq N_p\},
\end{equation} 
\noindent and $t_k = k \cdot \delta T$ 
denotes the timestamp of the $k^{th}$ predicted position $\xi_k$. Note that pure linear extrapolation may result in obstacle-colliding 
predictions. In such cases, a remedial primitive-based method from \cite{ji2022elastic, 
gao2023adaptive} is then applied, selecting collision-free primitives with 
minimal control effort as the target prediction. The predicted sequence $Q_{target}$ provides target information for the visibility-aware planning framework, which we detail in the subsequent sections.

\section{Spherical Signed Distance Field}
\label{sec:spherical_signed_distance_field}

We introduce a novel visibility-aware approach for target tracking that utilizes the SSDFs. Traditionally applied as a visibility model for spatial shading in computer graphics\cite{wang2009all, wang2014parallel, wang2012analytic}, SSDFs inherently align with visibility problems through their radial environmental representation centered at the focal point (the target). Lines of sight emanating from the target partition the space into visible and occluded sectors. At a specified tracking distance (radius), the degree of occlusion can be indicated by the signed spherical distance from the tracker's position to the closest visible sector boundaries. This section details our efficient SSDF updating method for real-time visibility-aware planning.

\subsection{Visibility Map Update}\label{sec:visibility_map_update}
\begin{figure}
    \begin{center}         \includegraphics[width=0.9\columnwidth]{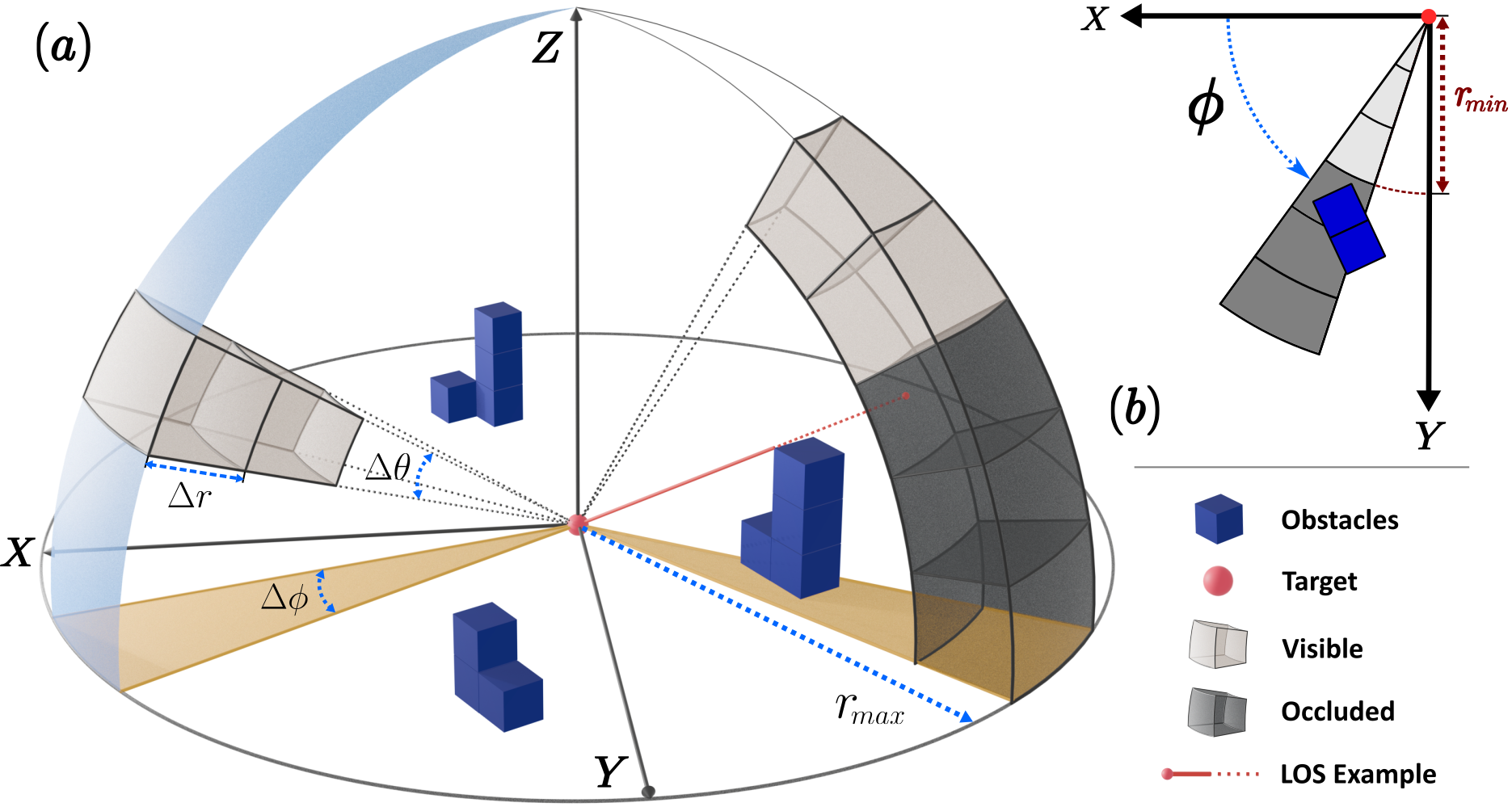}
    \end{center}
    \caption{\textbf{(a)} An illustration of the spherical discretization and the definition of the visibility map. A map cell is flagged as occluded if obstacles block the LOS. \textbf{(b)} An illustration of the monotonic property used in visibility maps. Along the direction $(0,\phi)$ in the figure, an obstacle blockage is at radius $r_{min}$. Then all grids with radii larger than $r_{min}$ can be directly set as occluded.}
    \label{fig:spherical_coordinate}
    \vspace{-0.1cm}
\end{figure}

To construct an SSDF around the target, we first propose to spherically parameterize the 3-D space and update a binary visibility map from the occupancy data. The visibility map, denoted as $\mathcal{V}$, is a 3-D grid where each cell has a binary state: a state of $1$ means it is visible, while $0$ signifies its occlusion from the target. The visibility map is discretized into a spherical grid defined by $\theta$ (polar/latitudinal angle), $\phi$ (azimuthal/longitudinal angle), and $r$ (radial distance), with the target position as the grid's origin. Here, $\theta_i$, $\phi_j$, and $r_k$ represent the respective indices of these parameters. Fig.~\ref{fig:spherical_coordinate}(a) shows the spherical grid. LOS visibility has a useful property: for a given direction, visibility deteriorates monotonically as the radial distance $r$ increases. If the LOS along the direction $(\theta, \phi)$ is firstly blocked by an obstacle located at $[\,\theta, \phi, r_{min}]_S$, then all positions $[\,\theta, \phi, r]_S$ with $r > r_{min}$ are also occluded. Fig.~\ref{fig:spherical_coordinate}(b) shows this property. We use $[\cdot]_S$ to denote the spherical coordinates. Exploiting this property, we update $\mathcal{V}$ by identifying the closest occluded radius $k_{min}$ for each direction $(\theta_i, \phi_j)$. In Alg.~\ref{alg:visibility_map_update}, we traverse the 
occupancy map $\mathcal{C}$ around the target within radius 
$r_{max}$ and record the smallest $k_{min}$ for every direction. All grids with radial index $k > k_{min}$ are directly set as occluded by \textbf{setOcclusion()}. The auxiliary array $S$ in Line 13 is prepared for 
the further updates in Sec.~\ref{sec:incremental_3-D_SSDF_update}.

\begin{algorithm}
\caption{Visibility Map Update} 
\label{alg:visibility_map_update}
    \begin{algorithmic}[1]
        \Require the grid map $\mathcal{C}$ and cells $\textbf{c}$ in $\mathcal{C}$; the visibility map $\mathcal{V}$ discretized by latitude $\theta_i$, longitude $\phi_j$, and radial distance $r_k$; the index of closest occluded radial distance $k_{min}[i,j]$ for direction $(\theta_i, \phi_j)$; The resolution $N_{r}$ of the radial dimension $r$; An auxiliary array $S$ of length $N_{r}$;
        \Ensure the visibility map $\mathcal{V}$ is updated;\\
       \textbf{Initialize:} $k_{min} \leftarrow (N_r - 1)$,
        \For{each $\textbf{c}$ in $\mathcal{C}$}
             \If{cell $\textbf{c}$ is occupied} 
                \State $[\theta_i, \phi_j, r_k]_S\leftarrow $\,\textbf{ToSphericalCoordinate}$(\textbf{c})$;
                \If{$k_{min}[i,j] >$ $k$} 
                    \State $k_{min}[i,j] \leftarrow k$\,;
                \EndIf
            \EndIf
        \EndFor
        \For{each direction $(\theta_i, \phi_j)$ in $\mathcal{V}$}
            \State $h \leftarrow k_{min}[i,j]$;
            \State \textbf{setOcclusion}($\mathcal{V}$, $h$);
            \State $S[h]$.PUSH($(\theta_i, \phi_j)$);
        \EndFor
    \end{algorithmic}
\end{algorithm}

\subsection{2-D Spherical Distance Transform}\label{sec:2-D_spherical_distance_transform}

In this section, we introduce how to update a 2-D SSDF from a 2-D visibility map parameterized by $\theta$ and $\phi$.
A 2-D visibility map layer $\mathcal{V}_r$ is extracted from the 3-D map $\mathcal{V}$ by fixing the radial dimension to a specific $r$, describing the visibility of all directions at radius $r$. Given two directions $\textbf{v}(\theta, \phi)$ and $\textbf{u}(\theta', \phi')$ on $\mathcal{V}_r$, their angular distance is given by the spherical law of cosines:
\begin{equation}\label{eqn:L_distance}
    \mathcal{L}\{\textbf{v}, \textbf{u}\} = \cos^{\text{-}1}(\,
    \cos\theta'\cos\theta + \sin\theta' \sin \theta \cos|\phi - \phi'|
    \,).
\end{equation}

\begin{figure}
    \begin{center}
         \includegraphics[width=0.9\columnwidth]{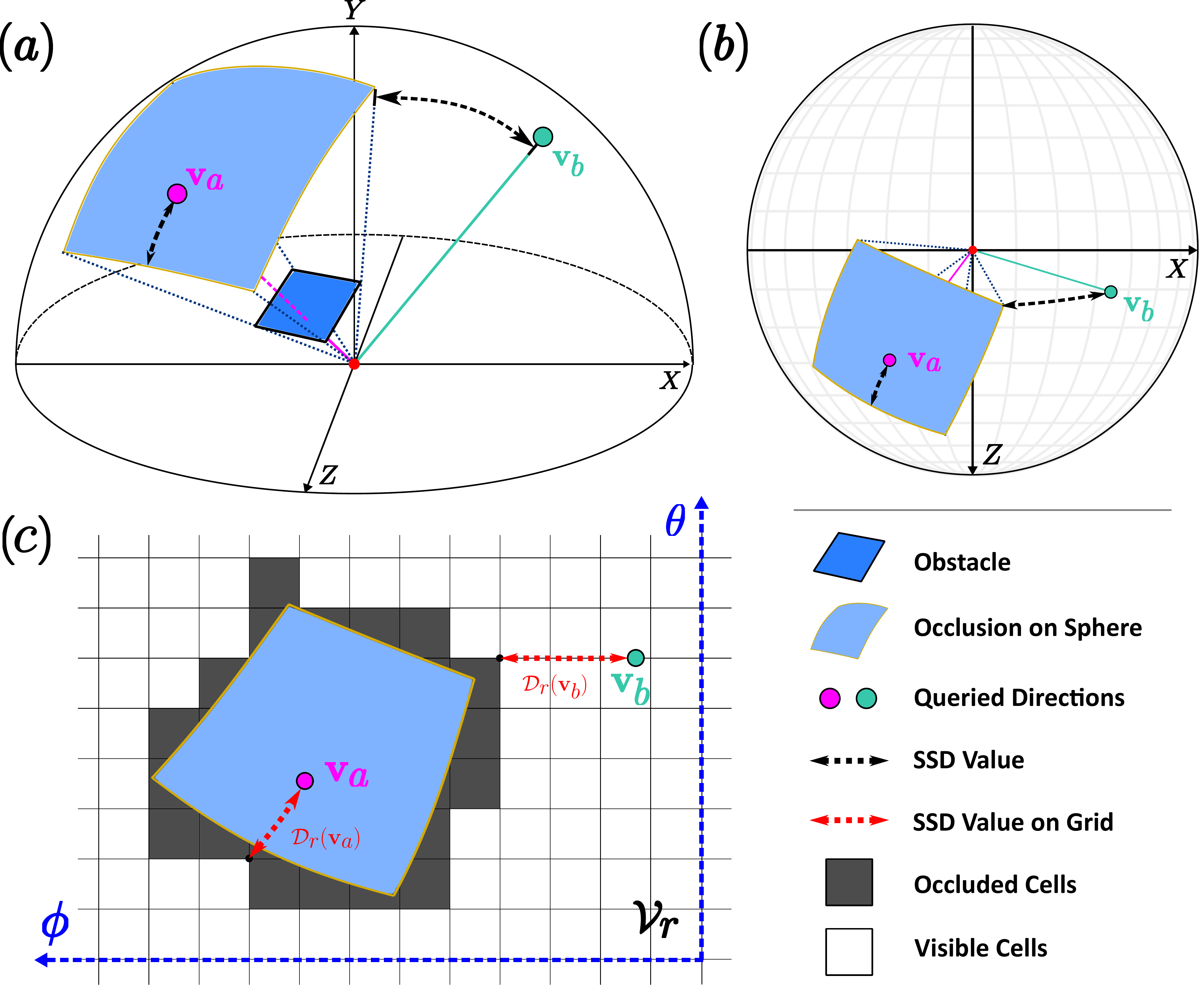}
    \end{center}
    \caption{Illustrations of the SSDF definition. \textbf{(a)} On the surface of a unit sphere, an occluded area is shadowed by an obstacle. For two queried directions $\textbf{v}_a$ and $\textbf{v}_b$, the black dashed curves on the sphere indicate the directions' Spherical Signed Distances (SSD) to the closest visibility boundary. \textbf{(b)} A top-down view of figure (a). \textbf{(c)} The corresponding occluded cells on the discretized 2-D $\theta$-$\phi$ grid. The red dashed curves indicate the SSDF values of the queried directions $\textbf{v}_a$ and $\textbf{v}_b$ on the discrete grid.}
    \label{fig:ssdf_def}
    \vspace{-0.22cm}
\end{figure}

A 2-D spherical distance field $\mathcal{D}_r$ is a more informative visibility model based on $\mathcal{V}_r$, where for a queried direction $\textbf{v}$, it stores a signed angular distance to the direction's closest visibility boundary. According to \cite{wang2009all, wang2014parallel, wang2012analytic}, it is defined as 
\begin{equation}\label{eqn:ssdf_definition}
    \mathcal{D}_r(\textbf{v}) = 
     \left\{\!\begin{array}{ll}
     \! +\min\limits_{\mathcal{V}_r(\textbf{u}) = 0} \mathcal{L}\{\textbf{u},\textbf{v}\},& \mathcal{V}_r(\textbf{v}) = 1, \\
    \! -\min\limits_{\mathcal{V}_r(\textbf{u}) = 1} \mathcal{L}\{\textbf{u},\textbf{v}\},&  \mathcal{V}_r(\textbf{v}) = 0,\\
    \end{array}\right.
\end{equation} 
where $\textbf{u}$ is a direction vector. By definition, SSDFs can quantify the degree of occlusion using the distance value, enabling trackers to query the minimum angular distance required to escape occluded regions.  Fig.~\ref{fig:ssdf_def}(a)-(b) illustrate the definition, and Fig.~\ref{fig:ssdf_def}(c) shows the 2-D $\theta$-$\phi$ grid for discrete distance transform. Unlike 2-D ESDFs~\cite{felzenszwalb2012distance, meijster2000general}, 
the non-Euclidean and asymmetric nature of the $\theta$-$\phi$ grid in SSDF prevents direct application of standard dimensionality-reduction algorithms. However, as shown in~\cite{wang2014parallel}, SSDFs can be correctly updated 
by scanning by reducing dimensions in a specific order: first latitudes, then longitudes.

We provide an overview of the algorithm in \cite{wang2014parallel} for clarity, which is a spherical version of \cite{felzenszwalb2012distance} following a certain scanning order. By definition in \eqref{eqn:ssdf_definition}, updating the spherical distance transform needs to find the closest visibility boundary for all the discretized directions $\textbf{v}(\theta_i, \phi_j)$. The process takes two phases.
In the first phase (Alg.~\ref{alg:latitudinal_scan}), for each latitude $\theta_i$, we find each direction's closest longitudinal visibility boundary $\phi_j^{cls}$ that minimizes the objective $\mathcal{L}\{(\theta_i, \phi_j),(\theta_i, \phi_j^{cls})\}$. This is a 1-D distance transform, which can be updated by a variant of the $L_1$-distance transform~\cite{felzenszwalb2012distance}. The array $g[i,j]$ in Alg.~\ref{alg:latitudinal_scan}
records the closest boundary direction $(\theta_i, \phi_j^{cls})$ 
for each $\phi_j$, initialized as $NULL$ (infinite distance point).

\begin{algorithm}
\caption{Latitudinal Scan}
\label{alg:latitudinal_scan}
    \begin{algorithmic}[1]
        \Require the visibility map $\mathcal{V}_r$ discretized by latitude $\theta_i$ and longitude $\phi_j$; The latitudinal resolution $N_{\theta}$ and the longitudinal resolution $N_{\phi}$; An auxiliary direction $h$;
        \Ensure the array $g$ is correctly updated;\\
       \textbf{Initialize:} $g \leftarrow NULL$,
        \For{$i = 0$ to $N_{\theta} - 1$} 
        \For{$j = 0$ to $N_{\phi} - 1$} 
            \If{$\mathcal{V}_r(\theta_i, \phi_j)$ is visible} 
                \State $g[i,j] \leftarrow (\theta_i, \phi_j)$;
            \Else
                \State $g[i,j] \leftarrow g[i,j-1]$;
            \EndIf
        \EndFor
        \State $h \leftarrow g[i,N_{\phi}-1]$;
        \For{$j = N_{\phi}-1$ to $0$} 
            \If{$\mathcal{L}\{g[i,j],(\theta_i, \phi_j)\} < \mathcal{L}\{h,(\theta_i, \phi_j)\}$}
                \State $h \leftarrow g[i,j]$;
            \EndIf
            \State $g[i,j] \leftarrow h$;
        \EndFor
        \EndFor
    \end{algorithmic}
\end{algorithm}

\begin{algorithm}
\caption{Longitudinal Scan}
\label{alg:longitudinal_scan}
    \begin{algorithmic}[1]
        \Require the visibility map $\mathcal{V}_r$ discretized by latitude $\theta_i$ and longitude $\phi_j$; The latitudinal resolution $N_{\theta}$ and the longitudinal resolution $N_{\phi}$; Two auxiliary arrays $v$ and $z$; 
        The array $B[i,j]$ recording closest visibility boundaries.
        \Ensure the closest boundary $D$ is correctly updated; 
       \For{$j = 0$ to $N_{\phi}-1$} 
       \State  $n \leftarrow 0$;
       \State  $v[0] \leftarrow 0$;
       \State  $z[0] \leftarrow -\infty$;
       \State  $z[1] \leftarrow +\infty$;
        \For{$i = 1$ to $N_{\theta}-1$} 
        \State $h$ $\leftarrow$ $\Theta(\,g[i, j],g[\,v[n], j], j)$;
        \While{$h\leq$ $z[n]$} 
        \State $n \leftarrow n - 1$;
        \State $h$ $\leftarrow$ $\Theta(\,g[i, j],g[\,v[n], j], j)$;
        \EndWhile
        \State $n \leftarrow n + 1$;
        \State $v[n] \leftarrow i$;
        \State $z[n] \leftarrow h$;
        \State $z[n+1] \leftarrow +\infty$;
        \EndFor
        \State$ n \leftarrow 0$
         \For{$i = 0$ to $N_{\theta}-1$}
         \While{$z[n+1]<i$}
         \State $n\leftarrow n+1$
         \EndWhile
         \State $B[i,j] \leftarrow 
         g[\,v[n], j]$;
         \EndFor
         \State \textbf{resetAuxiliaryArrays()};
        \EndFor
    \end{algorithmic}
\end{algorithm}

After deriving $\phi_j^{cls}$ in the first phase, the second phase updates the other dimension, $\theta_i$, to minimize the angular distance objective in \eqref{eqn:L_distance}. To achieve this, we conduct a variant of the $L_2$-distance transform scanning in \cite{felzenszwalb2012distance} along each longitude $\phi_j$, as detailed in Alg.~\ref{alg:longitudinal_scan}. While the distance objective $\mathcal{L}$ is not strictly an $L_2$-distance, it shares the same single intersection property \cite{wang2014parallel}, enabling the variant $L_2$-distance algorithm from \cite{felzenszwalb2012distance} to be applied. For two candidate boundaries $\textbf{v}_1(\theta_1, \phi_1)$ and $\textbf{v}_2(\theta_2, \phi_2)$ at a given longitude $\phi_0$, there exists a unique latitude $\theta_0$ where 
$\mathcal{L}\{\textbf{v}_1, \textbf{v}_0\} = \mathcal{L}\{\textbf{v}_2, \textbf{v}_0\}$. This single intersection latitude is computed by:

\begin{equation}\label{eqn:longitudinal_intersection}
    \Theta(\textbf{v}_1,\textbf{v}_2,\phi_0) = \left\{\!\begin{array}{ll}
     \!\pi/2,& P = Q, \\
    \! \tan^{\text{-}1}({\frac{R}{P-Q}}),&  \frac{R}{P-Q} \geq 0,\\
    \!\tan^{\text{-}1}({\frac{R}{P-Q}}) + \pi,& \frac{R}{P-Q} < 0, \\
\end{array}\right.
\end{equation}
where $P = \sin \theta_1 \cos (\phi_1 - \phi_0)$, $Q = \sin \theta_2 \cos (\phi_2 - \phi_0)$, and $R = \cos\theta_2 - \cos \theta_1$. In Alg.~\ref{alg:longitudinal_scan}, the function \textbf{resetAuxiliaryArrays()} clears the values in auxiliary 
arrays $v$ and $z$ and resets them for processing 
the subsequent longitude column. The array $B[i,j]$ records the final closest visibility 
boundary, and the spherical distance transform is 
$\mathcal{D}_r(\theta_i, \phi_j) = \mathcal{L}\{(\theta_i, \phi_j), 
B[i,j]\}$. For further procedure details, readers can refer 
to~\cite{wang2014parallel},~\cite{felzenszwalb2012distance}.
The $L_1$-then-$L_2$ order arises from the spherical law of 
cosines~\eqref{eqn:L_distance}: at a fixed latitude, the 
distance reduces to a 1-D function of $|\phi - \phi'|$ amenable 
to an $L_1$-style scan, while across latitudes it exhibits the 
single-intersection property requiring an $L_2$-style scan. 
Reversing this order breaks the algorithm's correctness. A complete derivation is provided in the 
supplementary material\footnote{\url{https://github.com/hku-mars/Swarm-Tracker/blob/master/documents/supplementary_materials.pdf}}\cite{swarmtracker_supp} Sec.~S-I.

During target tracking, trackers should maintain LOS visibility to the target. In visible regions where $\mathcal{V}_r = 1$, the LOS visibility constraints are already satisfied. Thus, in practice, we can solely focus on updating the distance fields of invisible areas. The updated spherical distances are then used to penalize the tracker's trajectories within these occluded regions, preventing visibility loss. The necessary angular clearance around the invisible area can be achieved by inflating the occluded grid when updating the visibility map $\mathcal{V}$. Thus, under this updating rule, the definition in \eqref{eqn:ssdf_definition} is reformulated as:
\begin{equation}\label{eqn:dr_definition}
    \mathcal{D}_r(\textbf{v}) = 
     \left\{\!\begin{array}{ll}
     \! 0,& \mathcal{V}_r(\textbf{v}) = 1, \\
    \! -\min\limits_{\mathcal{V}_r(\textbf{u}) = 1} \mathcal{L}\{\textbf{u},\textbf{v}\},&  \mathcal{V}_r(\textbf{v}) = 0.\\
    \end{array}\right.
\end{equation}
Using this definition, the update can be applied 
exclusively to the occluded areas, reducing the computation by skipping 
updates for visible directions where $\mathcal{V}_r = 1$. Although 
the spherical distance in (\ref{eqn:dr_definition}) is truncated compared to the original 
definition in (\ref{eqn:ssdf_definition}), we still refer to the updated distance field 
$\mathcal{D}_r$ as the SSDF to maintain consistency with the 
terminology in computer graphics literature. Throughout the 
remainder of this paper, SSDF refers to this truncated formulation.

\subsection{Incremental 3-D SSDF Update}\label{sec:incremental_3-D_SSDF_update}

In Sec.~\ref{sec:2-D_spherical_distance_transform}, we introduced the two-phase algorithm for updating the SSDF $\mathcal{D}_r$ of a 2-D visibility map $\mathcal{V}_r$, where $\mathcal{V}_r$ is a single layer extracted from the 3-D visibility map $\mathcal{V}$ at fixed radial distance $r$.  Our next goal is to update the complete SSDF $\mathcal{D}$ for the entire map $\mathcal{V}$, enabling the trackers to query the 3-D visibility by $\mathcal{D}(\textbf{p})$. $\mathcal{D}(\textbf{p})$ is defined by the value of the corresponding 2-D SSDF layer at $r_\textbf{p}$:
\begin{equation}\label{eqn:whole_d_definition}
    \mathcal{D}(\textbf{p}) = \mathcal{D}_{r_\textbf{p}}(\textbf{v}_{\textbf{p}}),
\end{equation}
where $\textbf{p} \in \mathbb{R}^3$ is the queried position, $r_\textbf{p}$ is the radial distance at position $\textbf{p}$, and $\textbf{v}_{\textbf{p}}$ is the direction vector at $\textbf{p}$. A brute-force updating approach is repetitively applying 
the two-phase algorithm to every radial layer $r_k$, which could be computationally 
expensive when the resolution of $r_k$ is high.

\begin{figure}
    \begin{center}
         \includegraphics[width=0.94\columnwidth]{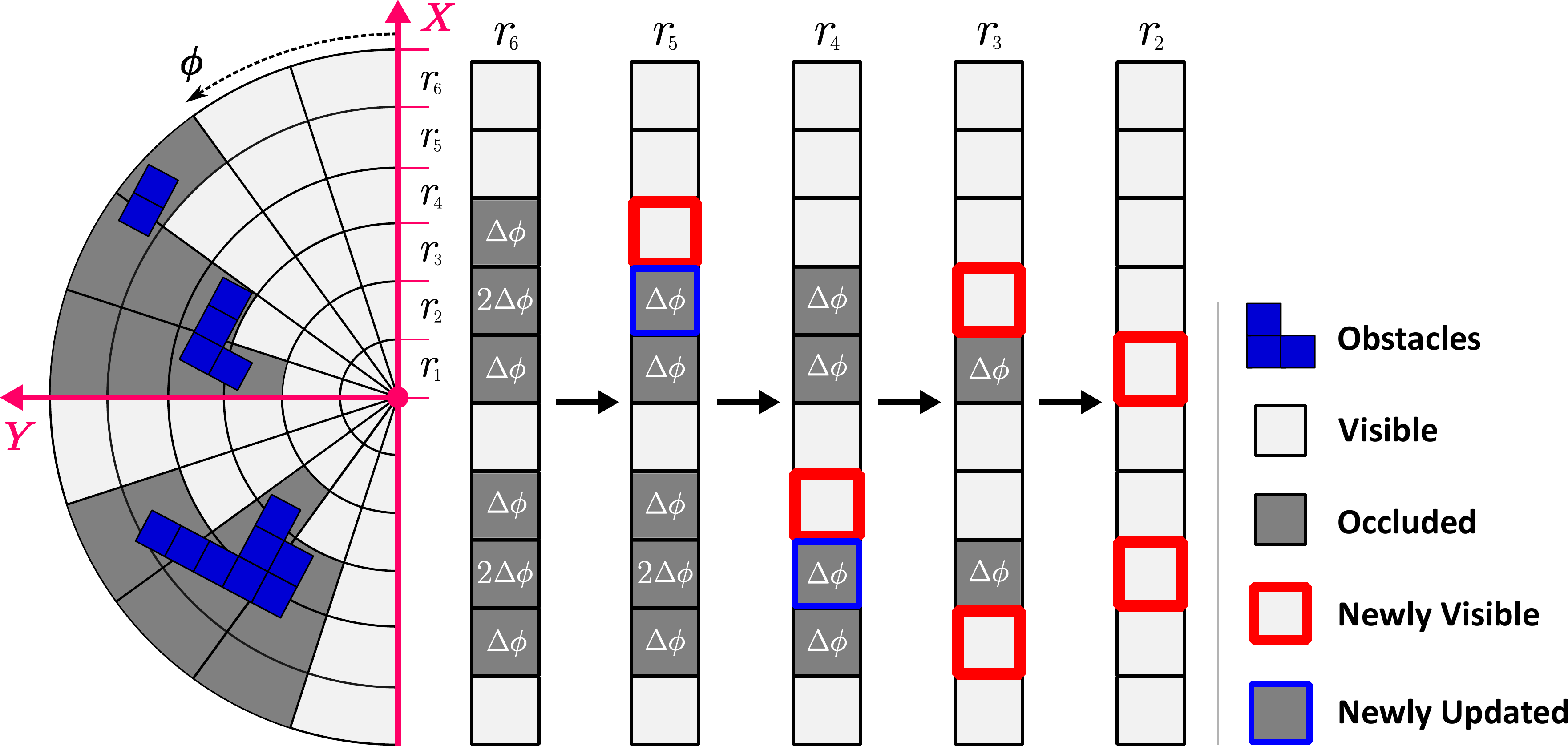}
    \end{center}
    \caption{An illustration of the incremental SSDF update strategy on a 2-D visibility map. The SSDF is updated layer by layer, starting from the outermost 1-D visibility layer at $r_6$ to the innermost layer at $r_1$. Proceeding to each layer, only the newly visible cells (in red boxes) and value-changed cells (in blue boxes) are identified and updated to compute the SSDF. In this example, only two cells in blue boxes require new value calculations throughout the update.}
    \label{fig:ssdf_incremental}
    \vspace{-0.25cm}
\end{figure}

Drawing inspiration from incremental updates for ESDFs \cite{han2019fiesta, oleynikova2017voxblox, pan2022voxfield, zhu2021vdb}, we instead propose an incremental strategy for 3-D SSDF computation leveraging the monotonic 
property from Sec.~\ref{sec:visibility_map_update}: along each direction $(\theta_i, \phi_j)$, visibility is non-increasing as the radial distance $r_k$ increases. So when $r_k$ decreases from $r_{max}$, each inner layer 
contains at least as many visible grids as the previous outer 
layer. We first update the outermost layer $\mathcal{V}_{r_{max}}$ 
at radial index $k = N_r - 1$ using the two-phase algorithm. For 
each subsequent inner layer, we only insert the newly visible grids 
into the SSDF of the previous layer, avoiding a full recomputation. 
This process continues sequentially until the innermost layer at 
$k=0$ is completed. Fig.~\ref{fig:ssdf_incremental} illustrates 
this incremental strategy.

\begin{figure}
    \begin{center}
         \includegraphics[width=0.9\columnwidth]{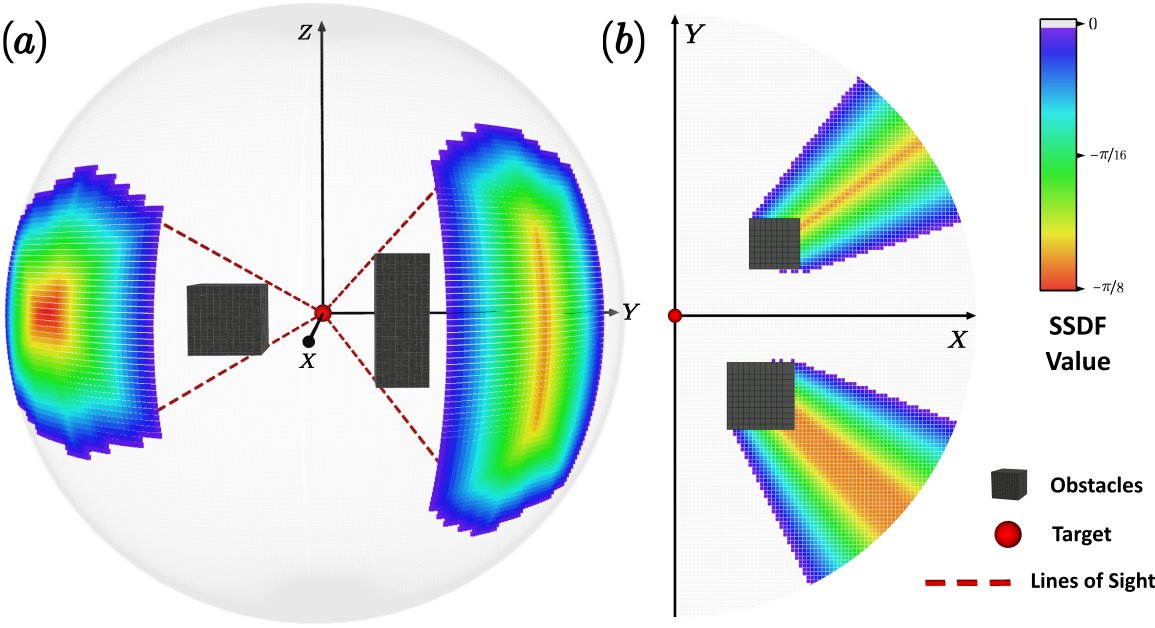}
    \end{center}
    \caption{An example of the SSDF computed for an occluded scene. \textbf{(a)} The SSDF values on the outermost spherical surface. \textbf{(b)} The horizontal cross-section of the updated SSDF at the $X$-$Y$ plane.}
    \label{fig:ssdf_result}
\end{figure}

We adopt the breadth-first-search (BFS) based incremental insertion 
from FIESTA~\cite{han2019fiesta} to implement this strategy. The 
newly visible directions at each layer have been collected in the 
auxiliary array $S$ from Alg.~\ref{alg:visibility_map_update}, 
where $S[k]$ contains all directions whose state changes from 
occluded to visible at the $k^{\text{th}}$ layer. In 
Alg.~\ref{alg:BFS}, each newly visible direction in $S[k]$ is 
initialized with zero distance and its own boundary, then BFS 
propagates updates to neighboring grids by comparing distances. 
Two arrays, $B[i,j]$ and $\mathcal{D}[i,j]$, record each grid's 
closest visibility boundary and spherical distance. By sequentially 
applying Alg.~\ref{alg:BFS} from the $(N_r-2)^{\text{th}}$ layer 
to the innermost one, the entire 3-D SSDF 
$\mathcal{D}(\textbf{p})$ is completed. The accuracy of this 
incremental approach is validated in 
Sec.~\ref{sec:Study on SSDF Updating Method}.
Fig.~\ref{fig:ssdf_result} shows an example of the updated SSDF.

\begin{algorithm}
\caption{Incremental SSDF Update}
\label{alg:BFS}
    \begin{algorithmic}[1]
        \Require $insertQueue$ as the queue for SSDF inserting; $k$ is the radial index of the current SSDF layer to update; The array $B[i,j]$ records grid $(\theta_i, \phi_j)$'s closest visibility boundary; The array $\mathcal{D}[i,j]$ records the spherical distance transform at $(\theta_i, \phi_j)$; Arrays $B$ and $\mathcal{D}$ are initialized by the results of the previous $(k+1)^{th}$ layer; 
        \Ensure the SSDF of the $k^{\text{th}}$ layer is updated;  \\
       \textbf{Initialize:} $insertQueue \leftarrow Empty$;
        \For{each newly visible direction $(\theta_i, \phi_j)$ in $S[k]$} 
        \State $B[i,j] \leftarrow (\theta_i, \phi_j)$;
        \State $\mathcal{D}[i,j] \leftarrow 0$;
        \State $insertQueue$.PUSH($(\theta_i, \phi_j)$);
         \EndFor
        \While {$insertQueue$ not empty}
         \State $(\theta_i, \phi_j) \leftarrow insertQueue$.FRONT();
         \State $insertQueue$.POP();
         \For{each neighbor $(\theta_m, \phi_n)$ of $(\theta_i, \phi_j)$}
        \If{$\mathcal{L}\{B[i, j], (\theta_m, \phi_n)\} < \mathcal{D}[m,n]$}
            \State $B[m, n] \leftarrow B[i, j]$;
            \State $\mathcal{D}[m,n] \leftarrow \mathcal{L}\{B[i, j], (\theta_m, \phi_n)\}$;
            \State $insertQueue$.PUSH($(\theta_m, \phi_n)$);
        \EndIf
        \EndFor
         \EndWhile
    \end{algorithmic}
\end{algorithm} 

\subsection{SSDFs Update on Target Prediction}\label{sec:SSDFs_for_target_tracking}

The last subsections outlined the procedures to update an SSDF around a target in 3-D spaces. To ensure SSDF-based visibility constraints throughout the tracking, we update SSDFs at every future position $\bm{\xi}_k$ of the predicted target sequence $Q_{target}$ in \eqref{eqn:q_target_set}. As the computations for these $N_p$ SSDFs are independent, they are updated in parallel, keeping the overall computation efficient. To enable gradient-based optimization with SSDFs, we employ the 
widely adopted the interpolation technique \cite{zhou2019robust, han2019fiesta} 
on the SSDF grid and compute the gradients analytically via 
differentiation of the trilinear interpolation formula. Although the spherical 
coordinate system is curvilinear, at sufficiently high resolutions this 
linear approximation remains accurate while maintaining 
computational efficiency. The application of SSDF for visibility-aware planning is detailed in Sec.~\ref{sec:visibility_aware_swarm_tracking}.

\textbf{Remark:} Ray-tracing provides binary visibility 
(visible or occluded), whereas SSDF provides a continuous 
spherical distance to the visibility boundary, paralleling the 
relationship between occupancy query and ESDF in obstacle 
avoidance. The SSDF's differentiability enables gradient-based 
optimization, which ray-tracing cannot support. In our pipeline, 
the visibility map (Sec.~IV-A) produces ray-tracing-equivalent 
binary states, upon which the SSDF is built. The proposed 
SSDF construction has a complexity of 
$O(N_\theta N_\phi + N_{change} \log N_{change})$, where 
$N_\theta$ and $N_\phi$ are the angular grid resolutions and 
$N_{change}$ is the number of cells whose visibility states 
change across radial layers. Detailed analysis is provided in 
the supplementary materials \cite{swarmtracker_supp} Sec.~S-II.

\section{Visibility-aware Cooperative Swarm Tracking}
\label{sec:visibility_aware_swarm_tracking}
This section introduces the cost functions modeling the swarm 
tracking requirements: avoiding environmental occlusions, 
maintaining tracking distance, keeping the target within the 
sensor's FOV, and coordinating trackers to prevent mutual 
occlusion while utilizing the surrounding space. These costs 
are used in both the front-end search and back-end optimization. In what follows, the ego drone (the $i^{th}$ in the swarm of $N$ drones) has 
position $\textbf{p} \in \mathbb{R}^3$, $\textbf{p}_j$ denotes 
the position of the $j^{th}$ teammate, and $\bm{\xi} \in 
\mathbb{R}^3$ denotes the target position.

\subsection{Environmental Occlusion Cost}\label{sec:environmental_occlusion_cost}
This cost $\mathcal{J}_{vis}$ is introduced to preserve the target's LOS visibility against static obstacle occlusions. The occlusion relationships in the environment w.r.t. the target are encoded by an SSDF introduced in Sec.~\ref{sec:spherical_signed_distance_field}. The environmental occlusion cost at position $\textbf{p}$ is designed as

\begin{equation}\label{eqn:ssdf_cost}
    \mathcal{J}_{vis} = - \mathcal{D}(\textbf{p}).
\end{equation}
As stated in Sec.~\ref{sec:spherical_signed_distance_field}, $\mathcal{D}(\textbf{p})$ returns the angular distance to the closest visibility boundary around $\textbf{p}$ if it is occluded, otherwise $\mathcal{D}(\textbf{p})$ returns zero. Both the front-end searching and back-end optimization use this cost to impose penalties on the occluded areas, thereby preventing visibility loss. To conduct numerical optimization with $\mathcal{J}_{vis}$, we derive the cost gradient as
\begin{equation}\label{eqn:ssdf_cost_grad}
    {\partial \mathcal{J}_{vis}}/{\partial \textbf{p}}=- \nabla \mathcal{D}(\textbf{p}),
\end{equation}
 where $\nabla \mathcal{D}(\textbf{p})$ is the interpolated gradient of SSDF at point $\textbf{p}$.

\subsection{Field-of-View Cost}\label{sec:field_of_view_cost}

The field-of-view (FOV) cost ensures that the target remains within 
the LiDAR's sensing range. Transforming the target position 
$^{w}\bm{\xi}$ from the world frame to the LiDAR frame yields:
\begin{equation}\label{eqn:field-of-view}
    ^{l}\bm{\xi} = \,_{w}^{b}\mathbf{R}(_{w}^{b}\textbf{q})\,(^{w}\bm{\xi} - \textbf{p}) - \,_{b}^{l}\textrm{\textbf{t}},
\end{equation}
where $_{w}^{b}\textbf{q}$ is the rotation quaternion, $_{w}^{b}\mathbf{R}$ 
is the corresponding rotation matrix, and $_{b}^{l}\textbf{t}$ is the 
LiDAR-body translation. In our system configuration, the rotation between the body frame and 
the LiDAR frame is identity.

\begin{figure}
    \begin{center}
         \includegraphics[width=0.955\columnwidth]{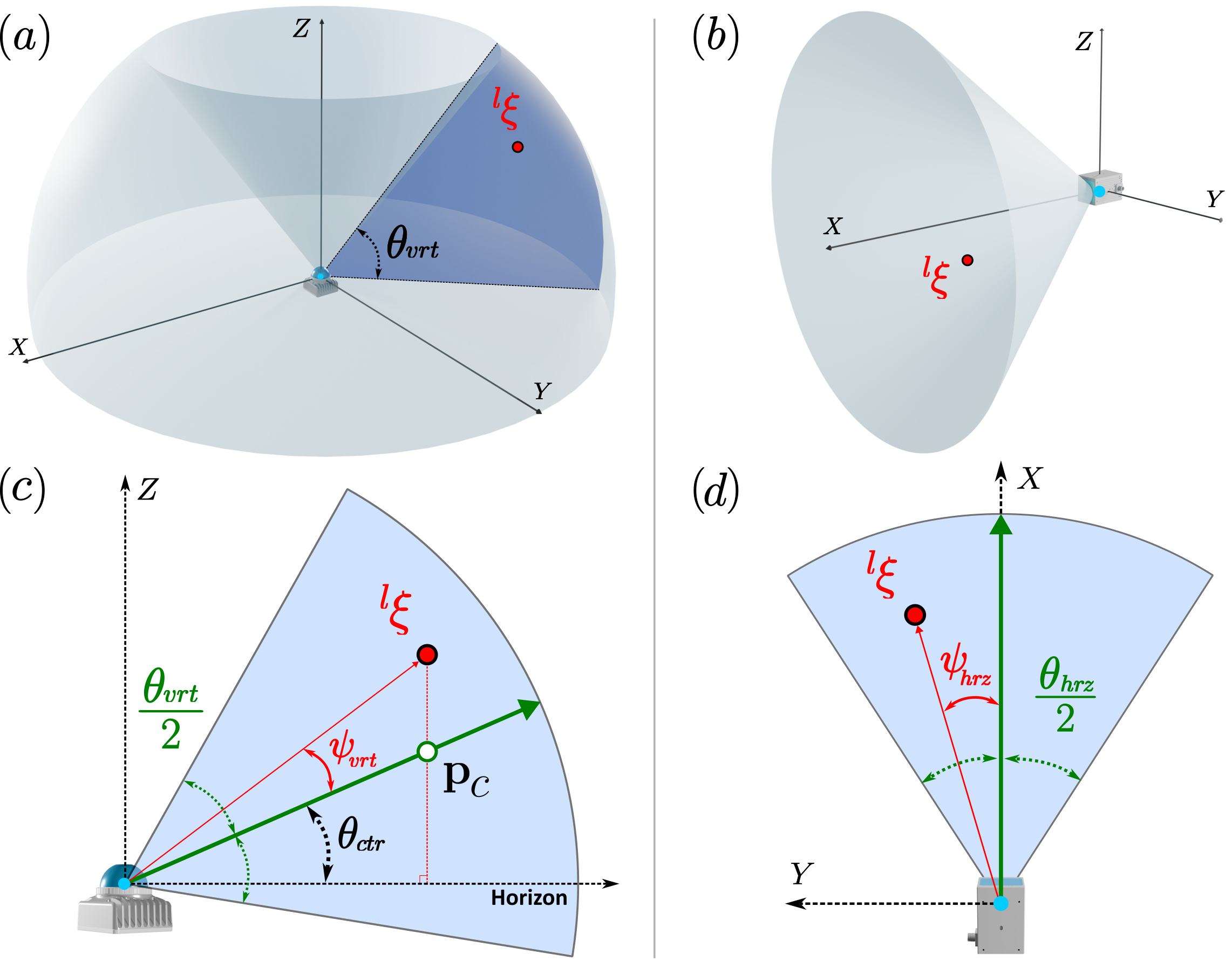}
    \end{center}
    \caption{\textbf{(a)} An overview of the Mid360 FOV. \textbf{(b)} An overview of the Avia FOV. \textbf{(c)} The vertical cross-section of the Mid360 FOV, with the angular bisector marked by the green arrow. To meet the FOV constraint, the angle $\psi_{vrt}$ should be less than $\theta_{vrt}/2$. \textbf{(d)} The horizontal cross-section of the Avia FOV, where the angle $\psi_{hrz}$ is expected to be zero to align with the target.}
    \label{fig:fov_cost}
\end{figure}

The cost formulation $\mathcal{J}_{fov}$ consists of vertical and horizontal parts. For sensors with horizontally omnidirectional but vertically limited FOV
(\textit{e.g.}, MID360), only the vertical FOV requires regulation. Given 
$^{l}\bm{\xi} = [x_l, y_l, z_l]^T$ and vertical FOV $\theta_{vrt}$, we 
define an auxiliary point on the vertical FOV bisector:
\begin{equation}
    \textbf{p}_c = [x_l, y_l, \sqrt{x_l^2 + y_l^2} \cdot \tan{\theta_{ctr}}],
\end{equation}
where $\theta_{ctr}$ is the angle between the vertical FOV bisector and the horizon level. As shown in Fig.\;\,\ref{fig:fov_cost}(c), to contain
the target in vertical FOV, the cost penalizes configurations where the angle $\psi_{vrt}$ between 
$^{l}\bm{\xi}$ and $\textbf{p}_c$ exceeds $\theta_{vrt}/2$:
\begin{equation}\label{eqn:vertical_fov_cost}
    \mathcal{J}_{fov}^{vrt} = \cos{\frac{\theta_{vrt}}{2}} - \frac{^{l}\bm{\xi}\cdot \textbf{p}_c}{\|^{l}\bm{\xi}\| \, \|\textbf{p}_c\|}.
\end{equation}

For sensors with conic FOVs (\textit{e.g.}, Avia), an additional horizontal 
cost should be imposed to align the drone's heading toward the target. The horizontal angle $\psi_{hrz}$ between the target and the heading axis in Fig.~\ref{fig:fov_cost}(d) is expected to be zero. Thus the horizontal cost can be given as
\begin{equation}\label{eqn:horizontal_fov_cost}
    \mathcal{J}_{fov}^{hrz} = 1 - \frac{x_l}{\sqrt{x_l^2 + y_l^2}}.
\end{equation}

Since $^{l}\bm{\xi}$ depends on both position $\textbf{p}$ and attitude 
$\textbf{q}$, $\mathcal{J}_{fov}$ is an $SE(3)$ cost. The gradients are:
\begin{equation}\label{eqn:fov_cost_gradient1}
    \frac{\partial \mathcal{J}_{fov}}{\partial \textbf{p}}
    =
    \frac{\partial \,\bm{^{l}\xi}}{\partial \textbf{p}} \frac{\partial \mathcal{J}_{fov}}{\partial \,\bm{^{l}\xi}},\;\; \frac{\partial \,\bm{^{l}\xi}}{\partial \textbf{p}} = - _{w}^{b}\mathbf{R}^T,
\end{equation}
\begin{equation}\label{eqn:fov_cost_gradient2}
    \frac{\partial \mathcal{J}_{fov}}{\partial \,_{w}^{b}\textbf{q}} 
    =
    \left(\frac{\partial \,\bm{^{l}\xi}}{\partial \,_{w}^{b}\textbf{q}} \frac{\partial \mathcal{J}_{fov}}{\partial \,\bm{^{l}\xi}}\right)^{-1},
\end{equation}
where $(\cdot)^{-1}$ denotes quaternion inversion.

\subsection{Tracking Distance Cost}\label{sec:tracking_distance_cost}
Let $d$ denote the Euclidean tracker-target distance, and 
$d_{lb}$, $d_{ub}$ be the desired lower and upper bounds. The 
cost is:
\begin{equation}\label{eqn:distance_cost}
    \mathcal{J}_{dis} = \left\{\begin{array}{ll}
 5\,(d_{lb} - d)^3,& d < d_{lb}, \\
 0,&  d_{lb} \leq d \leq d_{ub}, \\
 (d - d_{ub})^2 \,/ \,2,&  d > d_{ub},\\
\end{array}\right.
\end{equation}
where the coefficients (5 and 1/2) follow~\cite{yin2023decentralized}, 
enforcing a stronger lower-bound penalty for safety and a 
milder upper-bound penalty given the LiDAR's long sensing range. Unlike~\cite{zhou2022swarm, 
yin2023decentralized} that rigidly align the tracker and target 
altitudes for FOV compliance, our 3-D formulation, combined 
with the explicit FOV cost (Sec.~\ref{sec:field_of_view_cost}), 
allows flexible altitude adjustment.

\subsection{Teammate Occlusion Cost}\label{sec:teammate_occlusion_cost}
To prevent teammates from blocking each other's LOS to the target, 
trackers must maintain a minimum angular clearance $\theta_c$ 
relative to the target. Let $\eta_{ij} = \cos \angle(\textbf{p}_i - \bm{\xi},\, 
\textbf{p}_j - \bm{\xi})$ denote the cosine of the angular 
separation between teammate at $\textbf{p}_j$ and the ego drone 
at $\textbf{p}_i$, where $\angle$ denotes the angle between two 
vectors. When $\eta_{ij} > \cos\theta_c$, we impose an 
occlusion cost to repulse the tracker away from teammate $j$, 
and the total cost over all teammates is:
\begin{equation}\label{eqn:teammate_occlusion_cost}
    \mathcal{J}_{toc} = \sum_{j=1,j \neq i}^N\,
    (\eta_{ij} - \cos\theta_c)^3.
\end{equation}

\subsection{Swarm Distribution Cost}\label{sec:equidistant_distribution_cost}

Beyond maintaining the minimum angular clearance $\theta_c$, trackers 
should achieve uniform 3-D multidirectional target coverage. This 
equidistant distribution provides each tracker with maximum angular 
space to respond to occlusions or collisions, and maximizes the 
diversity of LiDAR measurement angles for more complete target point 
clouds. While the optimal angular spacing in 2-D is simply $2\pi/N$ 
for $N$ drones, the 3-D case is nontrivial. In this context, we introduce a new 
formulation for this 3-D scenario.

\begin{figure}
    \begin{center}
         \includegraphics[width=0.96\columnwidth]{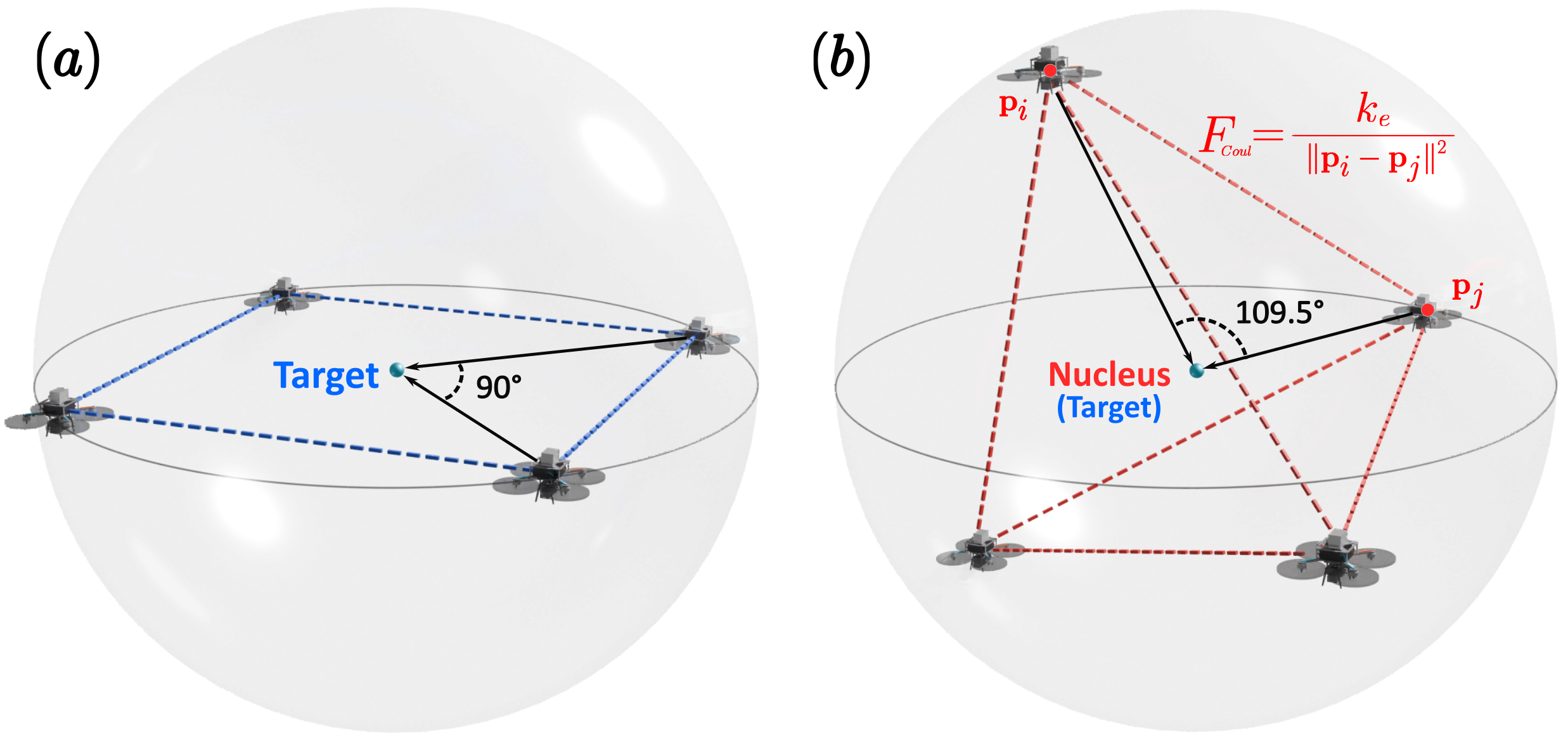}
    \end{center}
    \caption{Distributions of a four-drone swarm on a sphere centered at the target. \textbf{(a)} Conventional 2-D equidistant distribution that confines all drones in a plane. \textbf{(b)} The 3-D uniform encirclement distribution. Surrounding the target as a nucleus, the drones are modeled as electrons repelling each other with Coulomb's forces $F_{Coul}$, forming an optimal tetrahedral configuration to minimize total electrostatic potential energy.} 
    \label{fig:swarm_distribution}
    \vspace{-0.23cm}
\end{figure}

Inspired by the Thomson problem~\cite{tomson1904structure}, we 
notice that our desired equidistant target encirclement 
resembles its optimal configuration. The Thomson problem 
determines the minimum electrostatic potential energy 
configuration of $N$ electrons on a unit sphere repelling via 
Coulomb force, which naturally yields our desired distancing 
feature. Thus, we transform the distribution requirement into 
minimizing the swarm's total electrostatic potential energy. We adopt a logarithmic variant of the original Thomson problem as our cost formulation, which is from the 7th of the eighteen unsolved mathematics problems proposed by Steve Smale - "Distribution of points on the 2-sphere"\cite{Smale1998MathematicalPF}. The distribution cost for the $i^{th}$ drone is then designed as 
\begin{equation}\label{eqn:smale_cost}
    \mathcal{J}_{frm} = \sum_{{j=1,j \neq i}}^N k_e \log\frac{1}{\|\textbf{p}_i - \textbf{p}_j\|},
\end{equation}
where $k_e$ is an energy constant. A uniform multidirectional distribution could minimize this logarithmic potential energy objective. Fig.~\ref{fig:swarm_distribution} illustrates the model and compares the proposed 3-D distribution with the conventional planar formation. Compared to the 2-D square formation, the proposed distribution achieves a $109.5^{\circ}$ angle between teammate lines of sight, offering greater angular space for each tracker and increased diversity in viewing angles. This proposed cost describes the ideal distribution. However, in practical tracking, the FOV configurations in the swarm may not always allow for achieving the ideal distribution, but the planner strives to optimize the distribution cost within the FOV constraints.

\subsection{Other Costs}\label{sec:other_costs}

We employ safe flight corridors for obstacle avoidance, 
constraining each trajectory within polyhedral corridors via 
$\mathcal{J}_{obs} = \mathbf{A}_c \,\textbf{p} - b_c$, where 
$\mathbf{A}_c$ and $b_c$ are from the $\mathcal{H}$-representation. 
For dynamic feasibility, we limit the velocity, acceleration, 
and angular velocity amplitudes in the same form, 
\textit{e.g.}, $\mathcal{J}_{dyn}^{vel} = \|\textbf{v}\|^2 - 
v_{max}^2$. For inter-vehicle collision avoidance, each pair 
of drones maintains a distance clearance $r_s$ via the 
reciprocal cost
\begin{equation} 
\label{eqn:swarm_clearance}
    \mathcal{J}_{swm}^{j} = \max\{r_s^2 - \|\mathbf{E}^{1/2}
    (\mathbf{p}_i - \mathbf{p}_j)\|^2, 0\},
\end{equation}
where $\mathbf{E} = \text{diag}(1, 1, 1/c_e)$ with $c_e > 1$ 
defines an ellipsoidal metric~\cite{zhou2022swarm} enforcing 
larger vertical clearance to mitigate downwash. The total cost 
$\mathcal{J}_{swm}$ is summed over all teammates.

\section{Kinodynamic Searching}
\label{sec:kinodynamic_searching}

Our kinodynamic front-end generates a reference path by expanding motion primitives in a discretized control space. Unlike traditional hybrid A* searchers \cite{dolgov2008practical, liu2017search, zhou2019robust} that prioritize minimizing control effort along the path, our method scores each primitive based on the tracking performance metrics defined in the previous section, ensuring high consistency between the front-end and the overall task objectives. Readers unfamiliar with 
primitive-based motion planning are referred to  \cite{dolgov2008practical, liu2017search, zhou2019robust, yin2023decentralized}
for a comprehensive background.

\subsection{Primitive Expansion and Rejection}\label{sec:node_expansion}
\begin{figure}
    \begin{center}
         \includegraphics[width=0.96\columnwidth]{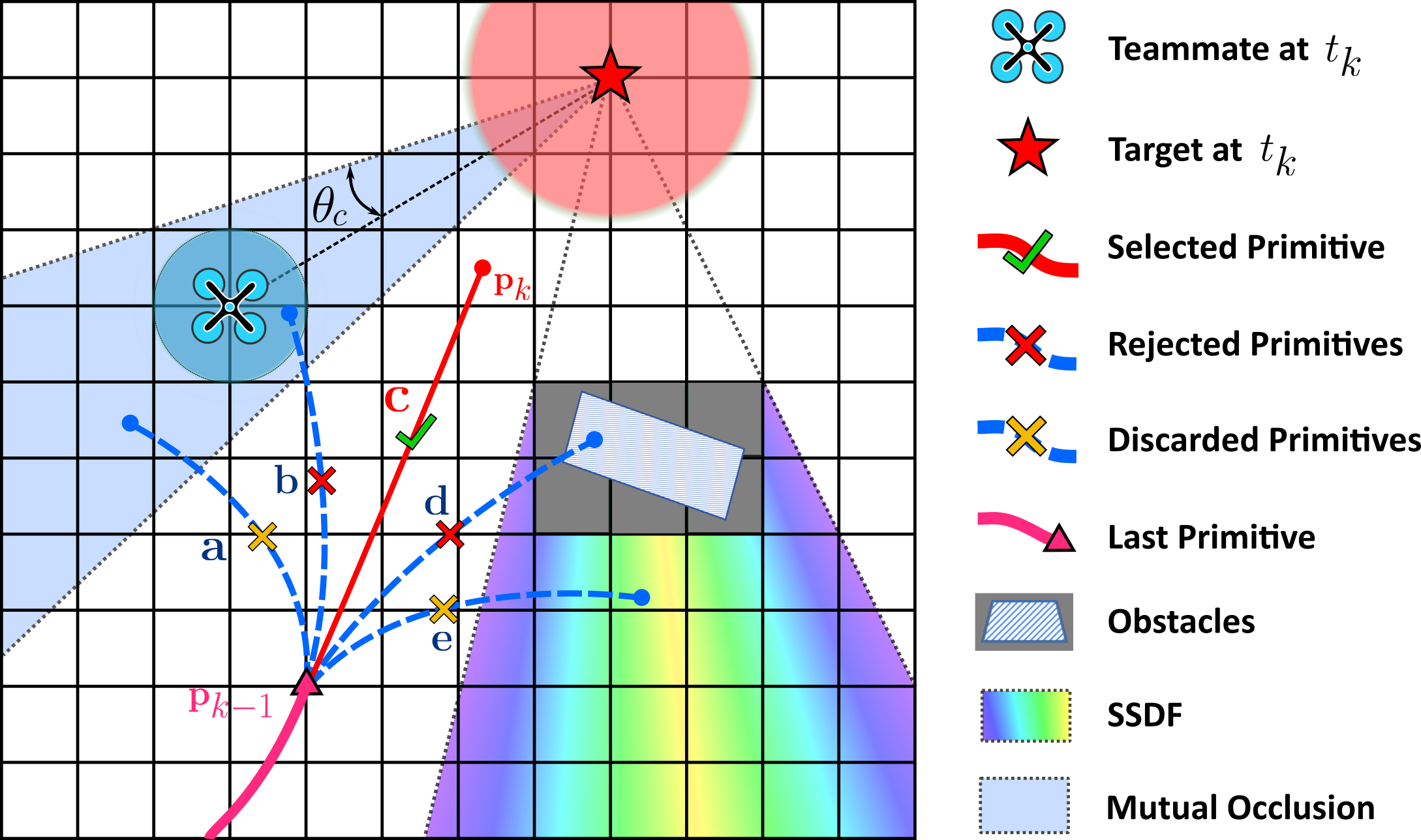}
    \end{center}
    \caption{An illustration of the primitive selection mechanism in our task-oriented kinodynamic searcher. Primitives $\textbf{b}$ and $\textbf{d}$ are rejected by inter-UAV safety check and obstacle check, respectively. Primitives $\textbf{a}$ and $\textbf{e}$ are penalized due to inter-UAV occlusion and environmental occlusion, respectively.  }
    \label{fig:kino_search}
\end{figure}

The state $\textbf{x} \in \mathbb{R}^6$ of the tracker drone includes its position $\textbf{p} = [p_x, p_y, p_z]^T$ and velocity $\textbf{v} = [v_x, v_y, v_z]^T$. Acceleration is used as the control input for each dimension, and the input space is discretized as $\textbf{u}_d = \{-a_{max}, 0, a_{max} \}$, where $a_{max}$ is the acceleration limit. This results in $3^3 = 27$ possible control inputs 
$\textbf{u}_d \in \mathbb{R}^3$ per expansion. In the front-end, motion primitives for the tracker drone are expanded directly using the prediction interval $\delta T$ of $ Q_{target}$ in (\ref{eqn:q_target_set}), ensuring that the timestamp $t_k$ of each new node $\textbf{x}_k$ aligns with the stamp of target prediction $\bm{\xi}_k$. The primitive expansion uses the double-integrator dynamics. After 
each expansion, nodes are pruned by checking obstacle 
avoidance, dynamic feasibility, and inter-vehicle safety. For 
inter-vehicle safety, a node is considered safe if its 
distances to all teammate positions (queried on broadcast 
trajectories) exceed a clearance $r_s$. 
Fig.~\ref{fig:kino_search} shows examples of primitive 
rejection.

\subsection{Cost Functions}\label{sec:frontend_cost_functions}
After the expansion and rejection process, every remaining node $\textbf{x}_k$ is assigned a cost $g_k$ as a coarse assessment of its tracking quality with target $\bm{\xi}_k$. We evaluate the performance of each node in terms of obstacle occlusion $\mathcal{J}_{vis}$, tracking distance $\mathcal{J}_{dis}$, teammate occlusion $\mathcal{J}_{toc}$, and swarm distribution $\mathcal{J}_{frm}$. Existing works~\cite{ho20213d, yin2023decentralized} penalize 
occlusion using voxel occupancy along the LOS, which is not a 
proper measure of occlusion severity. The severity should be 
characterized by the difficulty of escaping the invisible area, 
captured by the angular distance to the closest visibility 
boundary in SSDFs. We thus use $\mathcal{J}_{vis}$ to penalize 
occlusion. Synthesizing all the terms, we have the node cost $g_k$ as
\begin{equation}\label{eqn:front_end_costs}
    g_n = [\mathcal{J}_{vis},\, \mathcal{J}_{dis},\,\mathcal{J}_{toc},\, \mathcal{J}_{frm}] \cdot w,
\end{equation}
where $w$ is the weight vector. The searching terminates when one primitive reaches the target prediction horizon $T_p$, determining the final path. We utilize the remaining expansion time $h_n = T_p - t_k$ as a heuristic to speed up the searching process.
\subsection{Flight Corridor Generation}
\label{sec:corridor_generation}
After finding the minimum-cost path, an efficient method in \cite{liu2017planning} is applied to generate a safe flight corridor of connected polyhedra along the path, each polyhedron is denoted as
\begin{equation}\label{eqn:corridor}
    \mathcal{P} = \{x\in \mathbb{R}^3\; | \;\mathbf{A}_c\,x \leq b_c\}.
\end{equation}
The corridors will be used as the safe constraints. 

\section{Spatiotemporal Trajectory Optimization}
\label{sec:spatial_temporal_trajectory_optimization}

\subsection{Trajectory Optimization Problem Formulation}
\label{sec:trajectory_optimization_problem_formulation}

In this work, we adopt the MINCO trajectory representation \cite{wang2022geometrically}, 
a minimum control effort polynomial trajectory class, for spatiotemporal 
trajectory optimization. An $M$-piece trajectory $\textbf{p}(t)$ is 
compactly parameterized by intermediate waypoints 
$\bm{\varrho}=(\varrho_1,\cdots,\varrho_{M-1}) \in \mathbb{R}^{m(M-1)}$ 
and time allocation $\textbf{T}=(T_1,\cdots,T_M)^T \in \mathbb{R}_{>0}^M$, 
with total duration $T_{\Sigma} = \sum_{i=1}^{M} T_i$. Here $\textbf{p}(t)$ denotes the time-parameterized trajectory for the positional state. The polynomial coefficients $\textbf{c}$ are determined by the mapping 
$\textbf{c} = \mathcal{C}(\bm{\varrho},\textbf{T})$ from MINCO \cite{wang2022geometrically}. While the cost function is naturally defined over $\textbf{c}$, this 
mapping enables cost optimization over the compact parameters 
$(\bm{\varrho}, \textbf{T})$, with gradients computed via chain rule.

The MINCO-based trajectory optimization minimizes control effort while satisfying 
tracking constraints. Using the penalty method, the problem is formulated as:
\begin{equation}
\label{eqn:opt_problem}
    \min_{\bm{\varrho},\textbf{T}} \;
    \underbrace{\int_0^{T_{\Sigma}} \| \textbf{p}^{(s)}(t) \|^2 \,dt}_{\mathcal{J}_E}
    + \int_0^{T_{\Sigma}}\mathcal{J}_{\mathcal{G}}\,dt 
    + \sum_{t \in \mathcal{T}}\mathcal{J}_{\mathcal{H}},
\end{equation}
where $\mathcal{J}_E$ is the control effort objective, $\mathcal{J}_{\mathcal{G}}$ 
penalizes constraints that must hold continuously over the entire trajectory 
(\textit{e.g.}, obstacle avoidance, dynamic feasibility, inter-agent collision 
avoidance), and $\mathcal{J}_{\mathcal{H}}$ penalizes tracking-related constraints 
enforced at the target's predicted timestamps $\mathcal{T}$ (\textit{e.g.}, 
visibility, FOV alignment, swarm distribution). 

The total duration $T_{\Sigma}$ is fixed to the prediction 
horizon $T_p$ via variable substitution 
(Sec.~\ref{sec:temporal_constraint_elimination}), aligning the 
terminal state with the target's last predicted position. 
Unlike typical hierarchical tracking planners that fix the 
terminal position to the front-end path, we treat the terminal 
states (except velocity) as optimization variables, with 
tracking constraints applied via $\mathcal{J}_{\mathcal{H}}$ at 
$t = T_{\Sigma}$. In this work, we adopt $\mathfrak{T}_{MINCO}^{\,s=4}$ trajectories (\textit{i.e.}, 
7-degree polynomials) to represent the drone position. For cases requiring yaw 
planning, such as the Avia tracker, a $\mathfrak{T}_{MINCO}^{\,s = 2}$ trajectory 
(\textit{i.e.}, a 3-degree polynomial) is used to represent the drone's yaw angle. 
The position and yaw trajectories are jointly optimized upon the cost objectives 
in \eqref{eqn:opt_problem}. The positional trajectories are initialized using the 
path obtained from the front-end kinodynamic searching, the yaw angles are initially 
set to head toward the target positions, and the initial piece duration $T_i = T_p/M$.

\subsection{Continuous Relative-time Penalty}
\label{sec:continuous_relative_time_penalty}
Denote the continuous penalty as $\mathcal{J}_{C}$. The continuous penalty $\mathcal{J}_{C}$ 
enforces constraints evaluated along the entire trajectory. Using constraint 
transcription~\cite{teo1993new}, penalties are numerically integrated 
by sampling each trajectory piece with the trapezoidal rule:
\begin{equation} \label{eqn:continuous_penalty}
    \mathcal{J}_{C} = \sum_i^M \frac{T_i}{\kappa_i}\sum_{j=0}^{\kappa_i}
    \bar{\omega}_j \mathcal{J}_{\mathcal{G}}(t_j),
\end{equation}
where $t_j = (j/\kappa_i)T_i$ is the relative time on the $i^{th}$ piece 
and $\bar{\omega}_j$ follows the trapezoidal rule. The continuous penalties 
include obstacle avoidance $\mathcal{J}_{obs}$, dynamic feasibility 
$\mathcal{J}_{dyn}$, and swarm reciprocal clearance $\mathcal{J}_{swm}$:
\begin{equation}
    \mathcal{J}_{\mathcal{G}} = \lambda_{\mathcal{G}} \, 
    [\mathcal{J}_{obs}, \mathcal{J}_{dyn}, \mathcal{J}_{swm}]^T,
\end{equation}
where $\lambda_{\mathcal{G}}$ is the penalty weight vector. 
Among these, $\mathcal{J}_{obs}$ and $\mathcal{J}_{dyn}$ depend 
solely on the ego drone's states, with gradients following the 
chain rule on MINCO~\cite{wang2022geometrically}. The swarm 
clearance constraint $\mathcal{J}_{swm}^{\phi} (\textbf{p}_i(t_j), 
\textbf{p}_{\phi}(\tau))$ additionally involves teammate 
positions $\textbf{p}_{\phi}(\tau)$ queried at the absolute 
timestamp $\tau = \sum_{l=1}^{i-1}T_l + (j/\kappa_i)T_i$, which 
introduces extra gradient dependence on the preceding piece 
durations $T_l$ through the absolute-to-relative time mapping.

\subsection{Discrete Absolute-time Penalty}
\label{sec:discrete_absolute_time_penalty}
The discrete penalty $\mathcal{J}_{D}$ assesses tracking-related 
constraints at absolute timestamps $t_k$ aligned with predictions:
\begin{equation}
    \mathcal{J}_{D} = \sum_{k=1}^{N_p} \delta T \cdot \mathcal{J}_{\mathcal{H}}(t_k),
\end{equation}
where $\delta T$ and $N_p$ are the prediction interval and horizon. 
$\mathcal{J}_{\mathcal{H}}$ includes all tracking constraints from 
Sec.~\ref{sec:environmental_occlusion_cost}--\ref{sec:equidistant_distribution_cost}:
\begin{equation}
    \mathcal{J}_{\mathcal{H}} = \lambda_{\mathcal{H}} \, 
    [\mathcal{J}_{vis}, \mathcal{J}_{fov}, \mathcal{J}_{dis}, 
    \mathcal{J}_{toc}, \mathcal{J}_{frm}]^T.
\end{equation}
The single-UAV tracking constraints $\mathcal{J}_{vis}$, 
$\mathcal{J}_{fov}$ and $\mathcal{J}_{dis}$ depend on the ego 
state only. Assume that $t_k$ is located on the $i^{th}$ piece of the trajectory, then the corresponding relative time $t_r$ of $t_k$ on the $i^{th}$ piece becomes
$
\label{eqn: tk To t}
    t_r = t_k - \sum_{l=1}^{i-1}T_l,
$
where $T_l$ denotes the preceding piece duration. The 
formulation of $t$ here brings in gradient dependence on all 
$T_l$ with $1\leq l \leq i$. For the swarm constraints 
$\mathcal{J}_{toc}$ and $\mathcal{J}_{frm}$, both the ego and 
teammate trajectories are queried at the same fixed absolute 
time $t_k$, so the teammate positions remain constant 
throughout optimization and introduce no extra gradient terms.

\subsection{Temporal Constraint Elimination}
\label{sec:temporal_constraint_elimination}
We eliminate the 
fixed-duration equality constraint by substituting $\bm{\iota} 
= (\iota_1, \ldots, \iota_M) \in \mathbb{R}^M$ for the time 
allocation\cite{wang2022geometrically}:
\begin{equation}
T_i = \frac{e^{\iota_i}}{1 + \sum_{j=1}^{M-1} e^{\iota_j}} T_{p}, 
\; T_M = T_{p} - \sum_{j=1}^{M-1} T_j.
\end{equation}
With $\bm{\iota}$, the constraint $T_\Sigma = T_p$ is satisfied 
by default. Since $T_p$ is constant, the terminal penalty 
$\mathcal{J}_{\mathcal{H}}$ at $t = T_\Sigma$ has no temporal 
dependence, so its temporal gradients at $T_\Sigma$ vanish.

\textbf{Remark:} Our planner employs a penalty method with 
discretized front-end, which is suboptimal compared to 
global optimization approaches. This prioritizes real-time 
performance for onboard replanning on resource-constrained 
platforms.

\section{Simulation and Benchmark}
\label{sec:benchmark}
In this section, we conduct extensive comparative studies to validate the performance of our swarm tracking planner. All simulations in this section are run on an Intel Core i9-12900K CPU with an NVIDIA GeForce RTX 3070Ti GPU.

\subsection{General Benchmark of Swarm Tracking Planners}
To demonstrate the advantages of our method, benchmark comparisons are conducted against other cutting-edge swarm tracking works. The proposed planner is compared with Zhou's work\cite{zhou2022swarm}, Ho's work\cite{ho20213d}, and Yin's work\cite{yin2023decentralized}. Zhou \textit{et al.} [6] is included as a representative visibility-unaware baseline to demonstrate the importance of visibility-aware planning, as it utilizes a predefined constant leader-follower formation without considering target visibility. Ho's planner\cite{ho20213d} and Yin's planner\cite{yin2023decentralized} actively consider the target visibility. Ho \textit{et al.}\cite{ho20213d} uses a centralized dynamic programming to search the occlusion-free formation configuration for the swarm. Yin \textit{et al.}\cite{yin2023decentralized} build the planar visible sectors to avoid environmental occlusion. Planners are fine-tuned to their best performance.

\begin{figure}
    \begin{center}         \includegraphics[width=0.946\columnwidth]{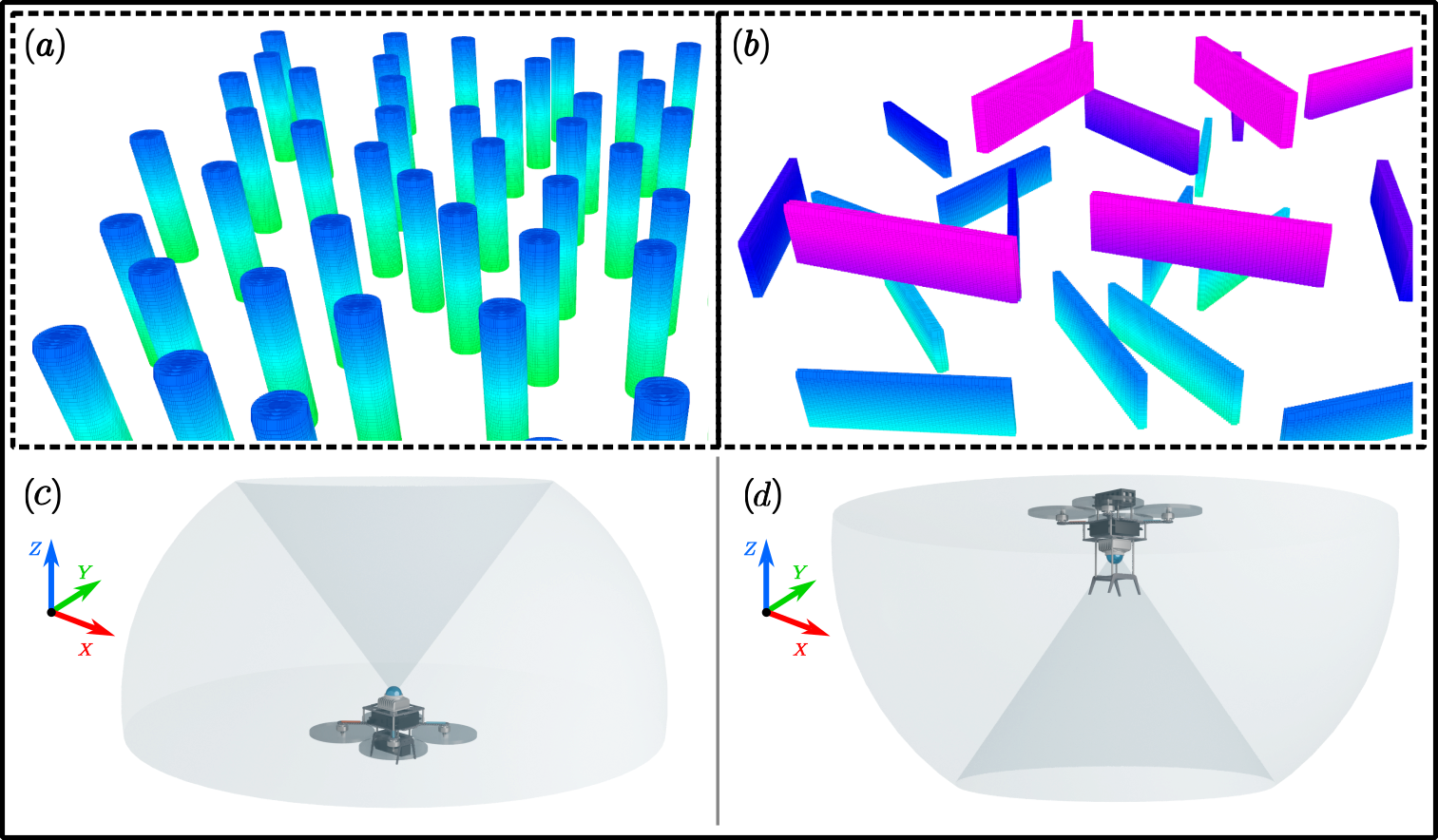}
    \end{center}
    \caption{\textbf{(a)} The Forest map used in the general benchmark. \textbf{(b)} The Walls map. \textbf{(c)} An illustration of the tracker drone with a regular upward-facing Mid360 LiDAR unit. \textbf{(d)} An illustration of the tracker drone with an inverted downward-facing Mid360 LiDAR unit.}
    \label{fig:benchmark_map_and_fov}
\end{figure}

To compare the planners fairly, we simulate four swarms, each with four trackers, chasing an identical target drone. Each swarm runs one benchmarked planner on two dense random maps: Forest (Fig.~\ref{fig:benchmark_map_and_fov}(a)) and Walls (Fig.~\ref{fig:benchmark_map_and_fov}(b)). Two target velocity 
modes are tested: slow ($1.0~m/s$) and fast ($2.5~m/s$). We also 
design two swarm configurations with different FOV settings: 
group A uses four upward-facing Mid360 LiDARs 
(Fig.~\ref{fig:benchmark_map_and_fov}(c)), while group B uses two upward-facing and two downward-facing Mid360 LiDARs (Fig.~\ref{fig:benchmark_map_and_fov}(d)). Testing is limited to horizontally omnidirectional FOVs due to the lack of yaw planning in other benchmarked planners.

The tracking performance is evaluated over four metrics 
from~\cite{yin2023decentralized}. A tracker is considered losing 
the target if the LOS is blocked by obstacles or teammates, the 
target exits the FOV, or the tracking distance becomes too close. 
The swarm visibility $\vartheta(t)$ counts the number of trackers 
not losing the target at time $t$. Based on this, we define: 
average visibility $\vartheta_{avg} = \frac{1}{T_b}\int_{0}^{T_b} 
\vartheta(t)dt$ over the task duration $T_b$, worst-case visibility 
$\vartheta_{wrst}$ as the minimum $\vartheta(t)$ over the task, 
full-visibility ratio $\gamma_{vis}$ as the fraction of time when 
all trackers observe the target, and average tracking distance 
$d_{avg}$ averaged in the same manner as $\vartheta_{avg}$. Each test case undergoes four independent trials, and the trial with the highest $\vartheta_{avg}$ is selected as the benchmark.
 
\begin{figure}
    \begin{center}         \includegraphics[width=1.0\columnwidth]{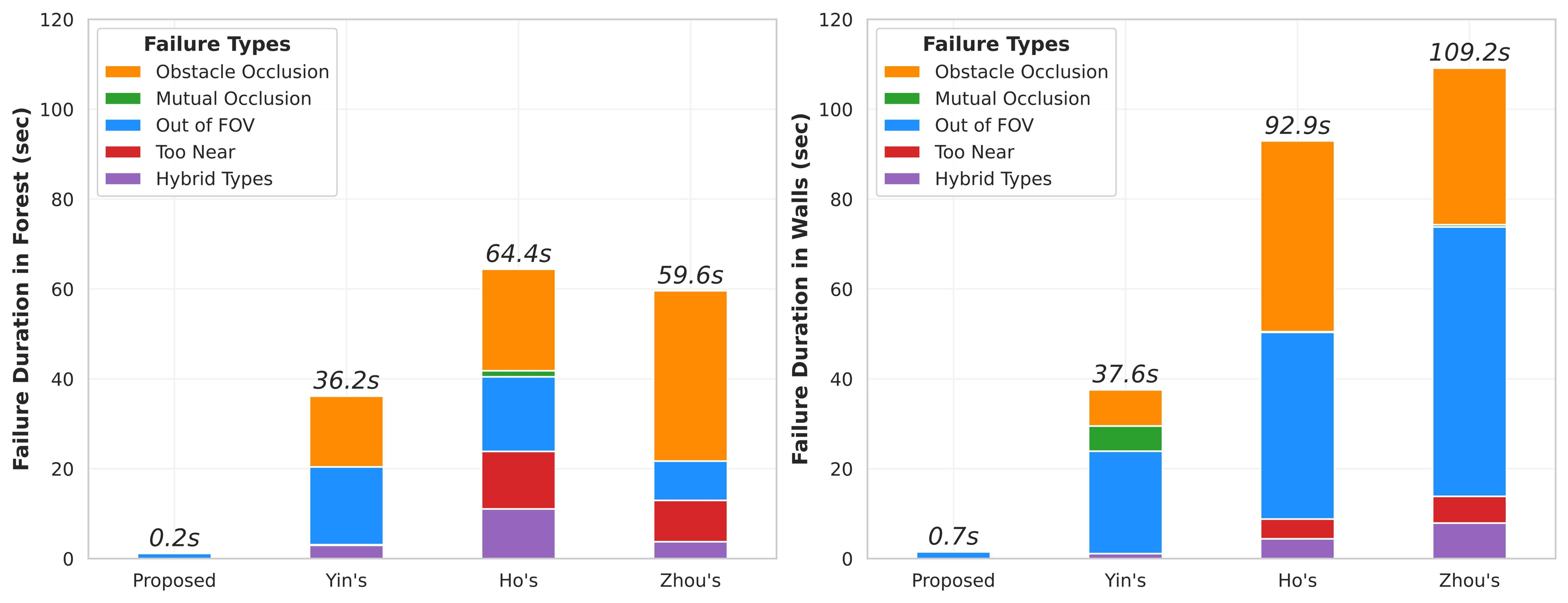}
    \end{center}
    \caption{Histograms of swarm-wide cumulative target loss duration with breakdowns by failure types (target velocity = $2.5 m/s$). \textbf{Left}: Histograms for the Forest map; \textbf{Right}: Histograms for the Walls map.}
    \label{fig:benchmark_histo}
\end{figure}

The results are summarized in Tab.~\ref{tab:general_benchmark_result}. Our method outperforms other works in terms of $\vartheta_{avg}$, $\vartheta_{wst}$, and $\gamma_{vis}$ in all cases with satisfactory distances $d_{avg}$. Notably, Zhou's and Ho's methods occasionally yield zero swarm visibility in worst-case scenarios, which could be fatal in real-world tracking applications. Significant performance degradation is observed in the three benchmarked planners when the target speed increases from $1.0~m/s$ to $2.5~m/s$. Fig.~\ref{fig:benchmark_histo} presents the cumulative target loss duration across all swarm trackers in the high-speed scenarios, along with its breakdown by specific failure reasons. The results in Tab.~\ref{tab:general_benchmark_result} also indicate worse tracking performance in the Walls environment compared to the Forest map, attributed to challenges such as difficult occlusion recovery from wall-shaped structures, increased vertical obstacle complexity, and greater target velocity fluctuations in the scenario. Importantly, our planner maintains near-complete swarm visibility ($>99\%$) across all test conditions, demonstrating its adaptability to the tested cluttered environments, varied FOV configurations, and different target velocities. 

\begin{table*}
    \caption{Visibility-aware Swarm Tracking Benchmark Results}
        \label{tab:general_benchmark_result}
        \centering
       
        \begin{tabularx}{\textwidth}{@{} 
        p{1.7cm} 
        >{\centering\arraybackslash}p{2.3cm}  
        *{4}{>{\centering\arraybackslash}X}  
        | 
        *{10}{>{\centering\arraybackslash}X} 
        c 
        @{}}
        \toprule
            & \multicolumn{1}{c}{\textbf{Configuration}} & \multicolumn{4}{c}{Swarm Configuration A} & \multicolumn{4}{c}{Swarm Configuration B}\\
        \toprule
        {\textbf{Scenario}}  & \diagbox[width=8em]{Method}{Metric} & $\bm{\vartheta}_{avg}$ & $\bm{\vartheta}_{wrst}$ & $\bm{\gamma}_{vis}$($\%$) & $d_{avg}$($m$) & $\bm{\vartheta}_{avg}$ & $\bm{\vartheta}_{wrst}$ & $\bm{\gamma}_{vis}$($\%$) & $d_{avg}$($m$)\\
        \toprule
        \multirow{4}{*}{\makecell[{{c}}]{\textbf{Forest}\\$(1.0\,m/s)$}}
        & Zhou \textit{et al.}\cite{zhou2022swarm} & 3.757 & 2.0 & 77.407 & 2.025 &  3.732 & 2.0  &  75.388 &  2.018  \\
        & Ho \textit{et al.}\cite{ho20213d} & 3.821 & 2.0 & 83.509 & 1.952 & 3.780  & 2.0  & 81.109  & 1.925   \\
        & Yin \textit{et al.}\cite{yin2023decentralized}  & 3.932 & 3.0 & 93.231 &  1.891  & 3.904 &  2.0  &  90.599 & 1.877   \\
        & \textbf{Proposed} & \textbf{4.000} & \textbf{4.0} & \textbf{100.000} & \textbf{2.016} &  \textbf{4.000} &  \textbf{4.0}  &  \textbf{100.000} &  \textbf{2.012}    \\
        \midrule
        \multirow{4}{*}{\makecell[{{c}}]{\textbf{Forest}\\$(2.5\,m/s)$}}
        & Zhou \textit{et al.}\cite{zhou2022swarm} & 3.535  & 0.0  & 62.323  & 2.052  & 3.538  & 0.0  & 60.786  &  2.035  \\
        & Ho \textit{et al.}\cite{ho20213d}  & 3.475  & 1.0 & 61.774  & 1.938 & 3.511  & 1.0 & 62.096  &  1.880   \\
        & Yin \textit{et al.}\cite{yin2023decentralized}  & 3.773 & 2.0 &  80.298 & \textbf{1.992} & 3.666  &   1.0  &  73.652  &  \textbf{2.006} \\
        & \textbf{Proposed}&  \textbf{3.997}  &  \textbf{3.0} &  \textbf{99.686} & 1.970 & \textbf{4.000}  & \textbf{4.0}  &  \textbf{100.000} &   1.980  \\
        \midrule
        \multirow{4}{*}{\makecell[{{c}}]{\textbf{Walls}\\$(1.0\,m/s)$}} 
        & Zhou \textit{et al.}\cite{zhou2022swarm}  & 3.433 & 1.0 & 64.613  & 2.030 & 3.430  & 1.0 & 61.069 & 2.020   \\
        & Ho \textit{et al.}\cite{ho20213d}  & 3.514  & 1.0 & 67.214  & \textbf{1.988} & 3.514  &1.0   & 67.214 & \textbf{1.988}  \\
        & Yin \textit{et al.}\cite{yin2023decentralized} & 3.919  & 3.0 & 91.917  & 1.925 & 3.874  & 2.0  & 88.274  & 1.905    \\
        & \textbf{Proposed} &  \textbf{4.000}  & \textbf{4.0}  & \textbf{100.000} & 1.926 &  \textbf{4.000} & \textbf{4.0} &  \textbf{100.000} &  1.976  \\
        \midrule
        \multirow{4}{*}{\makecell[{{c}}]{\textbf{Walls}\\$(2.5\,m/s)$}} 
        & Zhou \textit{et al.}\cite{zhou2022swarm}  & 3.182  & 0.0  &  53.383 & 2.072 & 3.194  & 1.0 &  49.590 & 2.101  \\
        & Ho \textit{et al.}\cite{ho20213d} & 3.252  & 1.0  & 50.037  & \textbf{1.955} & 3.365 &  0.0  & 58.451 & \textbf{1.962}  \\
        & Yin \textit{et al.}\cite{yin2023decentralized}  & 3.758  & 1.0 & 80.223  & 1.950  & 3.684 & 1.0 & 75.577  & 1.940    \\
        & \textbf{Proposed}  &  \textbf{3.993} & \textbf{3.0} & \textbf{99.331}  & 1.932 & \textbf{3.996}  & \textbf{3.0} & \textbf{99.628}  &  1.951   \\
       \bottomrule
       \end{tabularx} 
\end{table*}

\begin{figure*}[!t]
  \centering
  \includegraphics[width=0.96\textwidth]{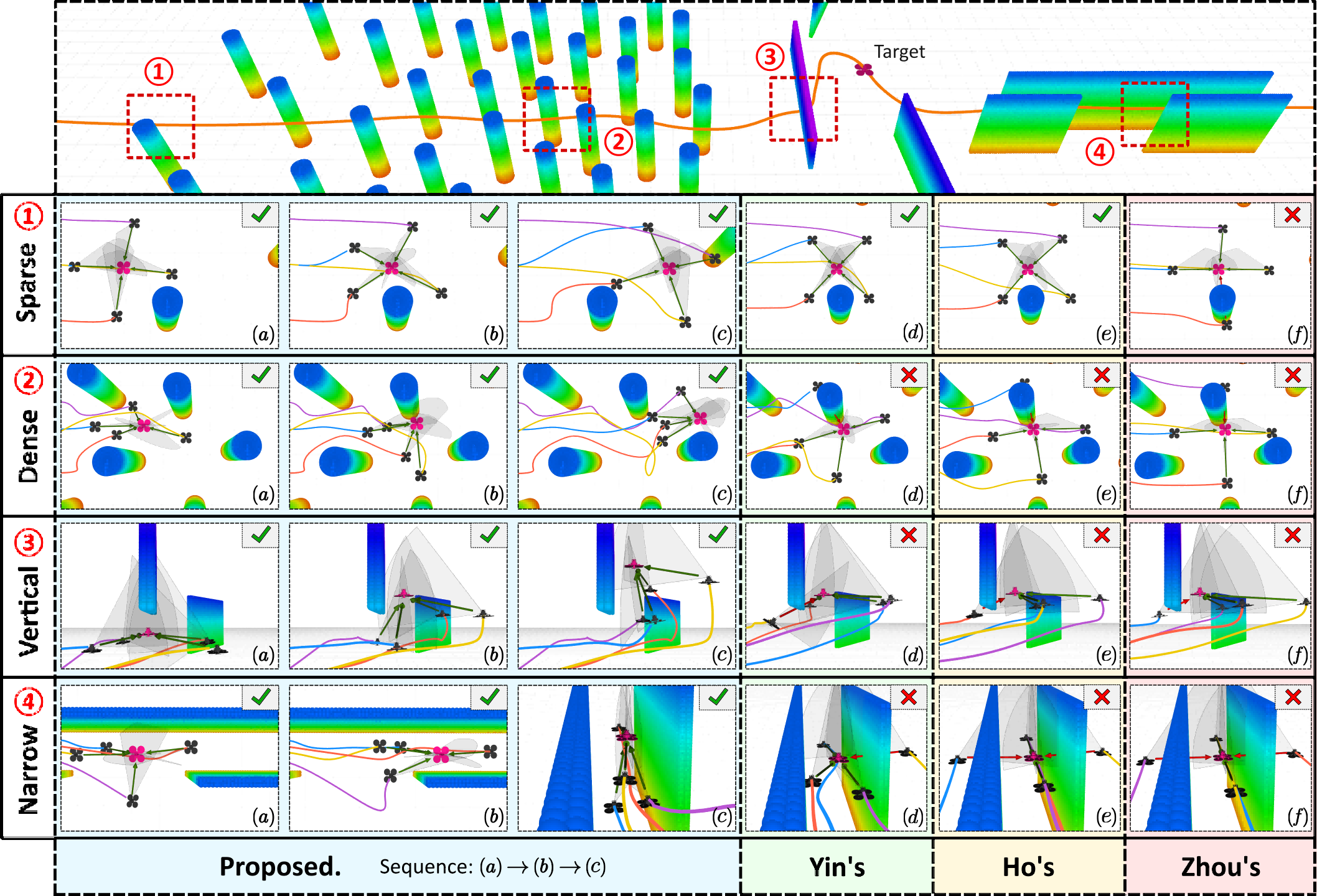}
  \caption{Tracking performance in the case study. \textbf{Top}: The orange trajectory is executed by the target. Four areas, \ding{172}-\ding{175}, are highlighted, featuring \textit{Sparse} clutter, \textit{Dense} clutter, \textit{Vertical} structures, and \textit{Narrow} passages. \textbf{Rows} \pmb{\ding{172}-\ding{175}}: Snapshots of swarm tracking in areas \ding{172}-\ding{175}. Subfigures (a)-(c) show the swarm behaviour of the proposed method. Benchmark results are shown in subfigures (d)-(f). The LiDAR FOVs' vertical cross-sections are depicted as gray sectors. }
  \label{fig:case_study}
\end{figure*}

\subsection{Case Study of Swarm Tracking Planners}
This section presents a case study comparing the methods. We establish 
a test map with diverse obstacle conditions (\textit{Sparse} clutter, 
\textit{Dense} clutter, \textit{Vertical} structures, and \textit{Narrow} 
passages). A four-drone swarm with upward-facing LiDARs tracks the target 
through the environment. Fig.~\ref{fig:case_study} illustrates the map 
and tracking behaviors, with green arrows indicating visible LOS and red 
arrows indicating visibility loss.

In the \textit{Sparse} area, Zhou's method loses visibility due to its 
simple constant leader-follower formation strategy, while other methods can avoid occlusion 
by adjusting the swarm. In the \textit{Dense} area, only our method 
maintains full visibility. Ho's method loses visibility as it cannot find occlusion-free rotation angles for its rigid square formation under dense clutter, and Yin's 2-D costs lack 
sufficient flexibility. For \textit{Vertical} obstacles, Yin's planner cannot 
timely avoid Z-axis occlusion since its visible sectors only consider 
2-D visibility at a fixed height. Our SSDF framework encodes visibility 
in full 3-D space, enabling prompt LOS adjustment against 3-D occlusion. 
In \textit{Narrow} passages, the other planners degrade due to the maneuverability limitations of 2-D motion 
constraints, whereas our method can coordinate the swarm at different heights, 
exploiting vertical space for full-visibility tracking with FOV compliance 
even in confined scenes.

\subsection{Ablation Study}
\label{sec:ablation_study}

We conduct ablation studies to validate the proposed modules 
and costs. The \textbf{w/o KinoSearch} variant replaces the 
kinodynamic front-end with a vanilla A* searcher 
from~\cite{ji2022elastic}. The \textbf{w/o Visibility} variant 
removes all SSDF-based occlusion costs, and the \textbf{w/o 
Formation} variant removes the 3-D swarm distribution costs. 
The \textbf{w/ voxel-occ} variant replaces the SSDF-based 
occlusion metric with the voxel-occupancy-based metric. Our 
full \textbf{Proposed} planner and Yin \textit{et al.}~[7] 
(the best-performing compared method) serve as the 
\textbf{Baseline}. All variants are tested in random Forest 
maps with five tree density levels from sparse 
($1/32~tree/m^2$) to dense ($1/9~tree/m^2$). Each cylindrical 
tree has a height of $4~m$ and a diameter of $1~m$, 
and the drone collision radius is set to $0.3~m$. Five 
trackers with upward-facing Mid360 LiDARs pursue a target 
moving up to $1.4~m/s$.

\begin{figure}
    \begin{center}         \includegraphics[width=0.9999\columnwidth]{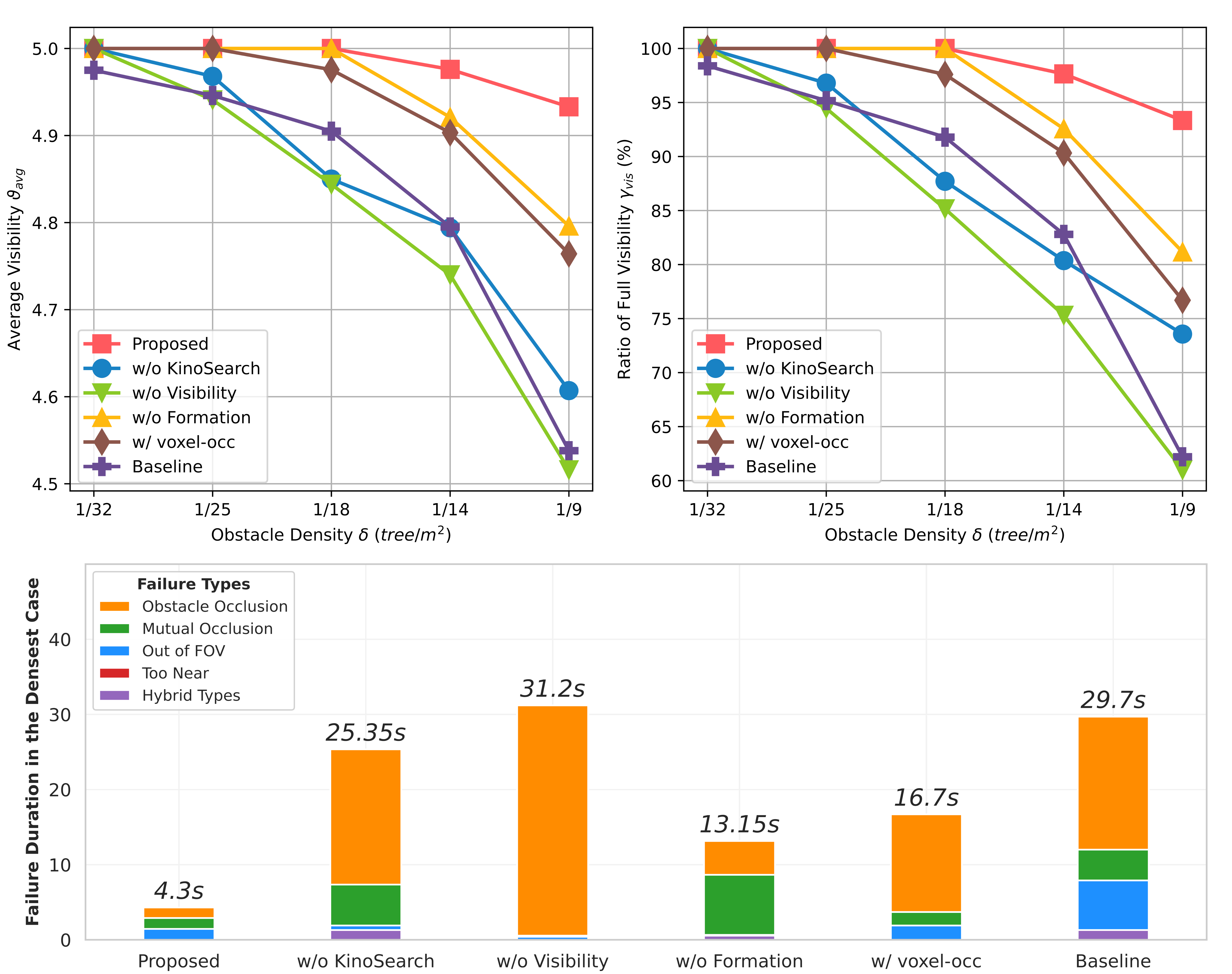}
    \end{center}
    \caption{Ablation study results. \textbf{Top-left}: The average visibility ($\vartheta_{avg}$) profiles across all variants. \textbf{Top-right}: The full visibility ratio ($\gamma_{vis}$) profiles across all variants. \textbf{Bottom}: The target loss duration histograms recording the test cases in the densest map ($1/9\,tree/m^2$).}
    \label{fig:ablation_charts}
\end{figure}

The ablation results are shown in Fig.~\ref{fig:ablation_charts}, which also depicts the visibility loss duration in the densest map. The A* searching in the variant \textbf{w/o KinoSearch} merely minimizes the path distance while disregarding other task requirements, thereby rendering low-performance path topologies that overburden the back-end optimizer. Without the information from SSDFs, the variant \textbf{w/o Visibility} becomes highly vulnerable to occlusions.  The variant \textbf{w/ voxel-occ} suffers from occlusion due to the metric limitation. While the variant \textbf{w/o Formation} leverages the kinodynamic searcher and SSDFs to reduce obstacle occlusions, the lack of swarm distribution guidance makes trackers prone to clustering behind the target in highly constrained scenes, inducing the risks of mutual occlusion. In contrast, the \textbf{Proposed} employs the distribution costs in both front-end and back-end to effectively mitigate the inter-vehicle interference, hence greatly enhancing the swarm visibility in dense areas. Through component ablation, we validate the contribution of the proposed modules and costs in ensuring the framework's overall performance. Using the same test settings, we further evaluate the 
tracking performance with limited-FOV Avia LiDARs. The results are provided in the supplementary materials \cite{swarmtracker_supp} Sec.~S-IV.

\subsection{Study on Swarm Size}
\begin{figure}
    \begin{center}         \includegraphics[width=0.96\columnwidth]{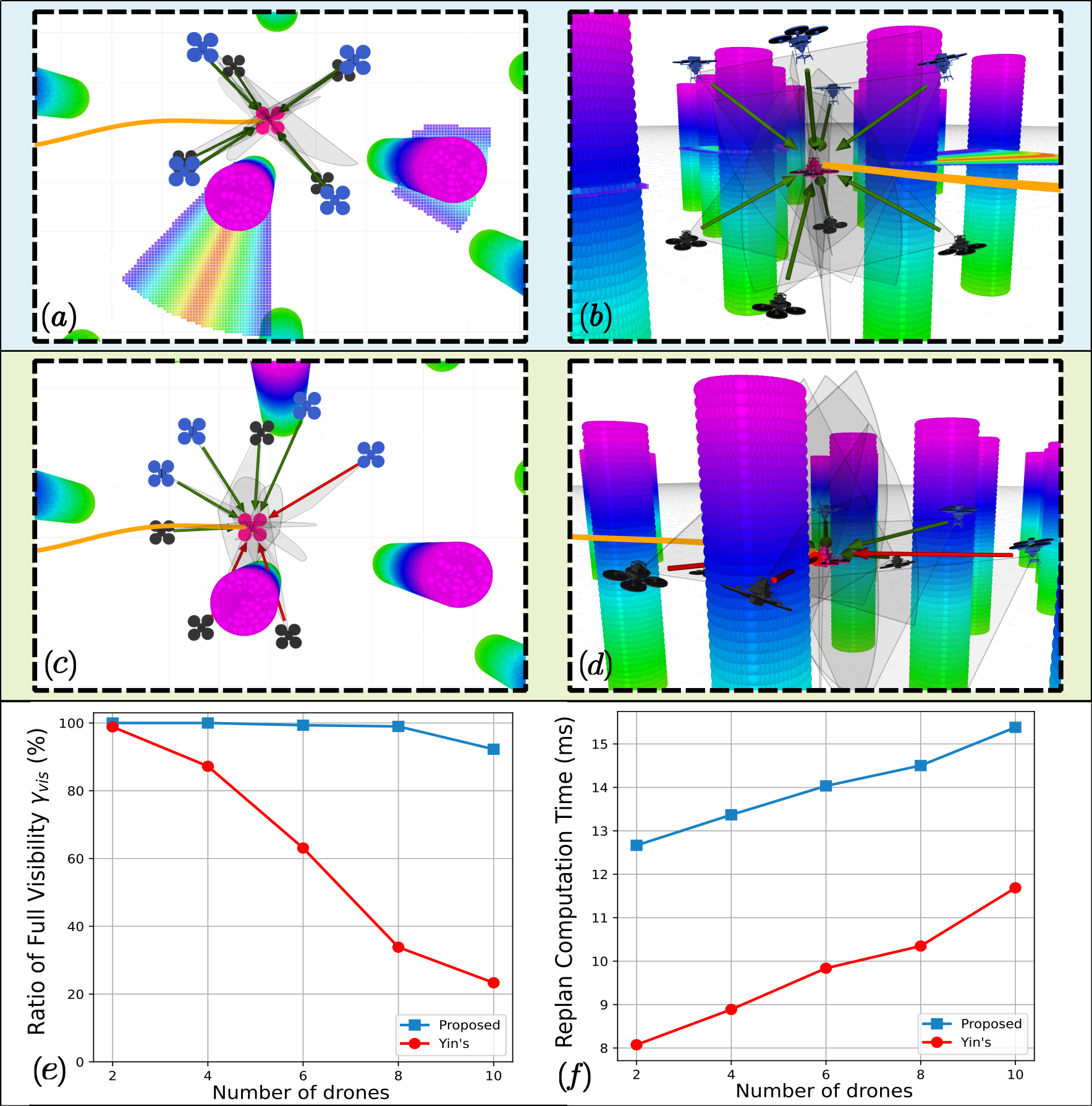}
    \end{center}
    \caption{Results of the swarm size study. \textbf{(a)-(b)} Top-down and side views of the proposed 8-drone tracking behavior, highlighting adaptive occlusion avoidance with a 3-D polyhedral swarm distribution. \textbf{(c)-(d)} Top-down and side views of the baseline tracking behavior, showing the failure to sustain full swarm visibility. \textbf{(e)} The full-visibility time ratio as the swarm scales up. \textbf{(f)} The computation time per replanning cycle as the swarm scales up.}
    \label{fig:scales_study}
\end{figure}

In this section, we conduct a scalability analysis of the proposed planner, investigating the impact of increasing swarm size on both tracking performance and computational overhead. Yin's work serves as the baseline. Five swarms of 2 to 10 drones are simulated, with half equipped with upward Mid360 LiDARs and half with downward ones. Fig.~\ref{fig:scales_study}(a)-(d) 
visualize the tracking behaviors of the eight-drone swarm, while 
(e)-(f) record the full-visibility ratio and average replanning time. As depicted in Fig.~\ref{fig:scales_study}(a)-(b), with the proposed FOV costs and spatial distribution costs, our planner can coordinate the swarm to exploit vertical airspace while ensuring strict FOV compliance. The drones adaptively configure into a 3-D polyhedral formation where downward-facing drones establish the upper facet and upward-facing drones form the lower facet. This distribution fully leverages the free space for occlusion-averse maneuvering. Conversely, Yin's method exhibits inefficient spatial utilization and visibility degradation as the swarm scales up, due to its 2-D motion constraints and lack of FOV modeling. Both methods maintain $<$$16~ms$ replanning time 
(Fig.~\ref{fig:scales_study}(f)) thanks to the decentralized 
architectures, with ours 4--5~$ms$ slower due to the $SE(3)$ 
back-end optimization. As the swarm size grows from 2 to 
10, our $\gamma_{vis}$ degrades by only $-0.97\%$ per drone, 
versus $-9.44\%$ per drone for Yin's method, demonstrating 
nearly $10\times$ better scalability.

\subsection{Study on SSDF Updating Method}
\label{sec:Study on SSDF Updating Method}
We compare the performance of our incremental SSDF update strategy (Sec.~\ref{sec:incremental_3-D_SSDF_update}) against the brute-force traversal approach. The brute-force traversal method iterates along the radial dimension in spherical coordinates ($\theta$-$\phi$-$r$), repetitively applying the 2-D SSDF updating process \cite{wang2014parallel} to each $\theta$-$\phi$ layer at every radial discretization $r$. As introduced in Sec.~\ref{sec:incremental_3-D_SSDF_update}, our incremental method strategically leverages the monotonic properties of occlusion across adjacent 2-D SSDF layers to reduce the overall update operations. 

\begin{figure}
    \begin{center}         \includegraphics[width=1.0\columnwidth]{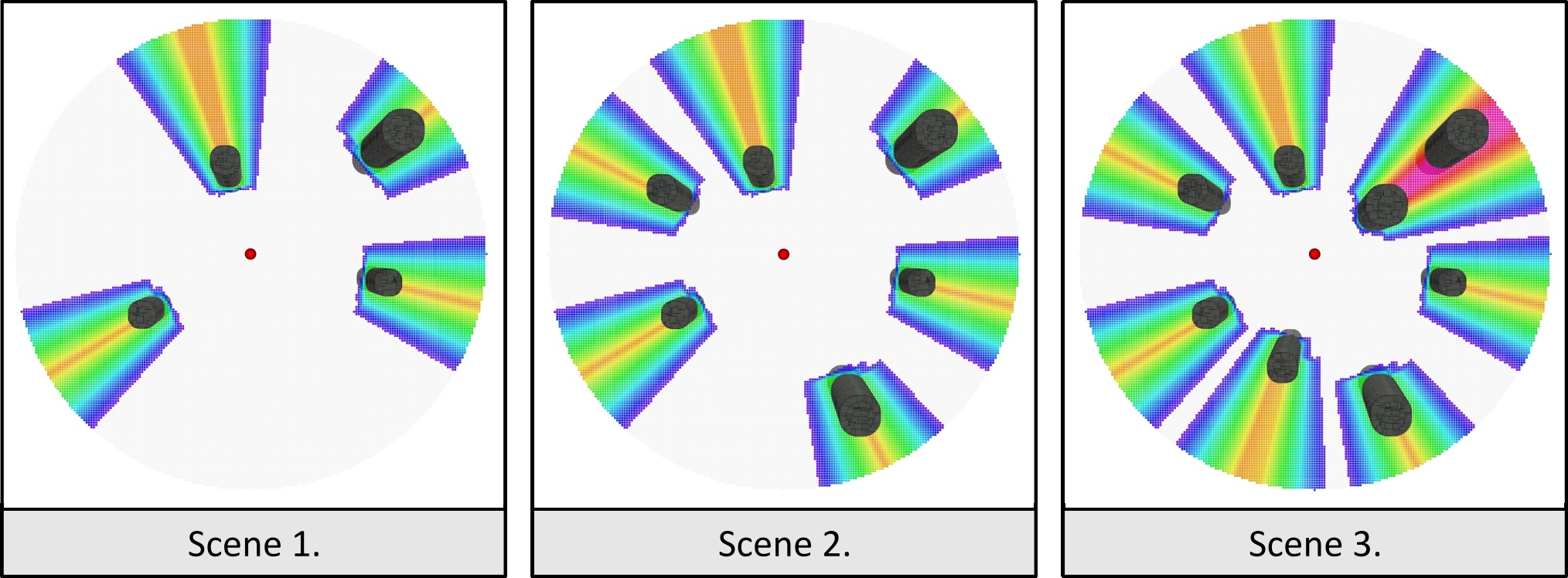}
    \end{center}
    \caption{Top-down views of the test scenes used in the SSDF updating study, along with cross-sections of the updated SSDFs at a height of zero.}
    \label{fig:benchmark_ssdf}
    \vspace{-0.35cm}
\end{figure}

We construct three scenes with increasing obstacle density, 
parameterizing the SSDF with a $5~m$ radial boundary, 0.1~m 
radial resolution, and 0.1~$rad$ angular resolution. 
Fig.~\ref{fig:benchmark_ssdf} shows the scenes and SSDF 
cross-sections. The brute-force method~\cite{wang2014parallel} 
provides ground truth values. As shown in 
Tab.~\ref{tab:ssdf_benchmark}, the proposed method achieves 
significantly faster update time ($<$4~$ms$ vs.\ $>$12~$ms$) with 
negligible cumulative error, confirming the efficiency and 
accuracy of the incremental strategy for real-time replanning.

\begin{table}[ht]
\centering
\caption{\scriptsize SSDF Update Benchmark Results
}
\label{tab:ssdf_benchmark}
\begin{tabular}{ccccc} 
\toprule
\diagbox[width=6em]{Scene}{Metric} & \makecell{Proposed Update\\Time ($ms$)} & \makecell{Baseline Update\\ Time ($ms$)} & \makecell{Cumulative Value\\ Error ($rad$)}  \\
\midrule
Scene 1 & 3.332 &  12.177 &  5.32 $\times 10^{-6}$ \\
Scene 2 & 3.688 &  12.230 &  8.31 $\times 10^{-6}$  \\
Scene 3 & 4.026 &  12.391 &  8.57 $\times 10^{-6}$ \\
\bottomrule
\end{tabular}
\end{table}

\subsection{Cost Weighting Ablation}
\label{sec:Cost Weighting Ablation}

The planner involves multiple cost terms combined through 
weight vectors $\lambda_{\mathcal{H}}$ and $\lambda_{\mathcal{G}}$ 
in Sec.~\ref{sec:spatial_temporal_trajectory_optimization}, 
following a hierarchical priority: safety and feasibility 
constraints ($\mathcal{J}_{obs}$, $\mathcal{J}_{dyn}$, 
$\mathcal{J}_{swm}$, $\mathcal{J}_{dis}$) have the highest 
weights; visibility constraints ($\mathcal{J}_{vis}$, 
$\mathcal{J}_{fov}$, $\mathcal{J}_{toc}$) have moderate weights; 
and the formation constraint ($\mathcal{J}_{frm}$) has the 
lowest weight. To assess weight sensitivity, we scale three key 
tracking cost terms ($\mathcal{J}_{fov}$, $\mathcal{J}_{vis}$, 
$\mathcal{J}_{frm}$) by factors of $\times 0.01$, $\times 0.1$, 
$\times 0.5$, and $\times 2$ relative to the default 
($\times 1$), tested in the $1/14$~$tree/m^2$ scene from 
Sec.~\ref{sec:ablation_study}. The default achieves 
$\vartheta_{avg} = 4.97$ and $\gamma_{vis} = 97.6\%$. As shown 
in Table~\ref{table:weight_ablation}, reducing the weights 
below the default leads to degradation, while 
increasing them yields marginal improvements, indicating 
a well-balanced default.
\begin{table}[h]
\centering
    \renewcommand\arraystretch{1.3}
    \caption{Cost Weighting Ablation Results 
    ($\vartheta_{avg}$ / $\gamma_{vis}$)}
    \label{table:weight_ablation}
    \scalebox{0.89}{
    \begin{tabular}{c !{\vrule} c c c c}
    \toprule
      \multirow{2}*{\makecell[c]{Cost \\ Term}} 
        & \multicolumn{4}{c}{Weight Scale} \\
      \cline{2-5}
        & $\times 0.01$ & $\times 0.1$ & $\times 0.5$ & $\times 2$ \\
       \hline
      FOV 
        & 4.74 / 80.3\% & 4.86 / 86.2\% & 4.92 / 92.0\% & 4.95 / 95.9\% \\
       \hline
      SSDF 
        & 4.83 / 83.6\% & 4.87 / 88.8\% & 4.94 / 94.4\% & 4.98 / 98.1\% \\
       \hline
      Formation 
        & 4.90 / 91.9\% & 4.87 / 92.4\% & 4.95 / 95.3\% & 4.97 / 97.3\% \\
       \hline
    \end{tabular}
    }
\end{table}

\section{Real world experiments}
\label{sec:experiment}

\begin{figure*}[!t]
  \centering
  \includegraphics[width=0.99\textwidth]{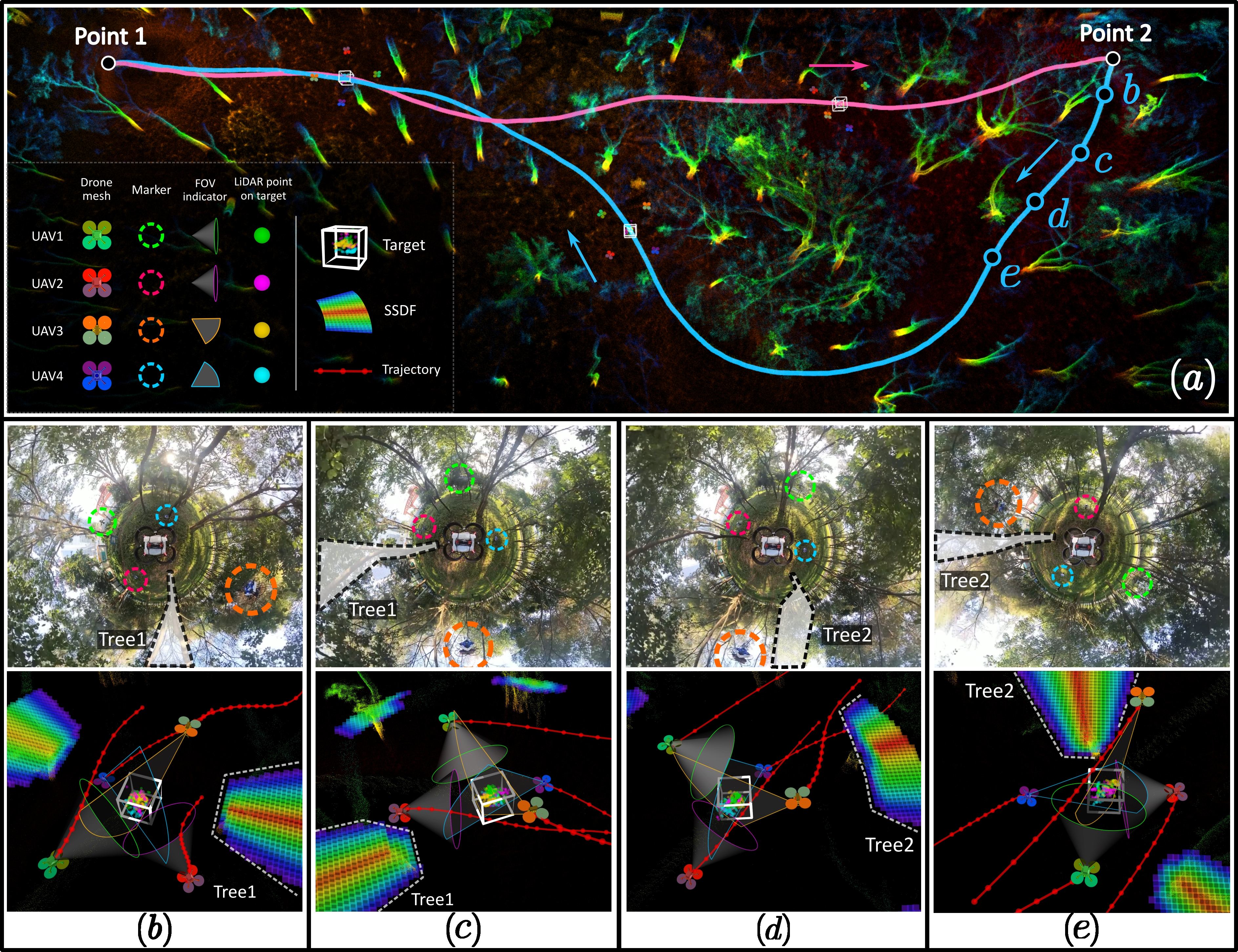}
  \caption{Four drones cooperatively track a flying target drone (up to 3~m/s) in a forest. \textbf{(a)} The pink curve depicts the target route from \textit{Point 1} to \textit{Point 2}, and the blue curve shows the return route. Four keyframes from $b$ to $e$ are selected for further illustration. \textbf{(b)-(e)} Each keyframe includes a snapshot with a 360° camera image (top) and an RViz visualization (bottom). Two tree obstacles, labeled \textit{Tree1} and \textit{Tree2}, along with their corresponding SSDFs, are highlighted to demonstrate occlusion avoidance. The heterogeneous drone FOVs are also visualized to illustrate the FOV compliance.}
  \label{fig:RR_exp}
\end{figure*}

\subsection{System Setup and Implementation Details}
\label{sec:System Setup and Implementation Detail}
We integrate the proposed framework with a decentralized LiDAR-based aerial swarm system. The swarm system is composed of autonomous drones with heterogeneous LiDAR configurations, including upward-facing Mid360 LiDAR, downward-facing Mid360 LiDAR, and Avia LiDAR. Each drone is equipped with a PixHawk flight controller and an onboard computer, the Intel NUC with an i7-12700 CPU. The entire swarm is localized using a decentralized swarm LiDAR-inertial odometry (Swarm-LIO) \cite{zhu2024swarm}, which provides 100~Hz state estimation and 25~Hz point clouds for each tracker. All drones are controlled by an on-manifold MPC\cite{lu2022manifold}. The drones communicate via a UDP wireless network. The targets are passive, exchanging no information with the 
trackers. Target measurement and state estimation follow 
Sec.~\ref{sec:target state estimation}, with high-reflectivity 
markers attached to the targets for detection. Target 
observations are shared among teammates via UDP for 
decentralized fusion. ROG-Map~\cite{ren2024rog} is used for 
occupancy grid mapping. For limited-FOV sensors (\textit{e.g.}, 
Avia LiDARs), we employ a bandwidth-efficient map 
synchronization framework~\cite{shi2024real} that encodes newly 
updated voxels into compressed chunks shared via UDP. Receivers 
spatiotemporally align and merge the data into their local 
maps, allowing Avia-equipped drones to maintain environmental 
awareness beyond their native FOV while orienting sensors 
toward the target. Our planner solves $SE(3)$ trajectory 
optimization using LBFGS-lite~\cite{wang2022geometrically} at 
15~Hz, with planned trajectories immediately shared for 
coordination. All modules run onboard in real time. The video link\footnote{\url{https://www.youtube.com/watch?v=lTPE_JnsTPI}}
shows the real-world experiments.

\subsection{Swarm Tracking in Dense Forest}
\label{sec:Swarm Tracking in Dense Forest}

To validate the real-world performance of our method, we test the tracking system in an unknown dense forest. A heterogeneous LiDAR-based swarm (one upward-facing Mid360, one downward-facing Mid360, and two Avia LiDARs) is deployed. The swarm tracks a manually triggered target drone flying through the forest with a velocity up to $3~m/s$. The prediction horizon of the target motion is $1.8~s$. The target flies from \textit{Point 1} to \textit{Point 2} in 
Fig.~\ref{fig:RR_exp}(a) and returns via a distinct route. 
Throughout the flight, our decentralized swarm performs 
visibility-aware tracking without collisions or occlusions, 
recorded by an Insta360 camera on the target drone. Four 
snapshots (Figs.~\ref{fig:RR_exp}(b)-(e)) illustrate how the 
trackers adjust the swarm distribution to avoid LOS blockages 
by \textit{Tree1} and \textit{Tree2}, leveraging the SSDFs for 
occlusion avoidance. Driven by the joint costs, the swarm 
self-organizes into a tetrahedral distribution while ensuring 
FOV compliance. Tab.~\ref{tab:exp_computation_time} summarizes 
the average onboard computation time for each stage: SSDF 
update ($t_{\text{SSDF}}$), front-end searching 
($t_{\text{search}}$), corridor generation ($t_{\text{SFC}}$), 
and back-end optimization ($t_{\text{optimize}}$). Drones with 
Avia LiDARs exhibit longer runtime due to the 
joint yaw trajectory optimization.

\begin{table}[ht]
\centering
\caption{\scriptsize Average Onboard Computation Time (milliseconds)
}
\label{tab:exp_computation_time}
\begin{tabular}{c|ccccc} 
\toprule
 & $t_{\text{SSDF}}$ & $t_{\text{search}}$ & $t_{\text{SFC}}$ & $t_{\text{optimize}}$ & $t_{\text{total}}$ \\
\midrule
Upward Mid360   & 7.76 &  0.26  &  3.17 &  9.35 &  20.54 \\
Downward Mid360 & 7.61 &  0.22  &  3.10 &  8.93 &  19.86 \\
AVIA            & 8.53 &  0.35  &  3.19 & 13.20 &  25.27 \\
\bottomrule
\end{tabular}
\end{table}

\subsection{Cooperative Human Runner Tracking}
\label{sec:Cooperative Human Runner Tracking}
\begin{figure*}[!t]
  \centering
  \includegraphics[width=0.99\textwidth]{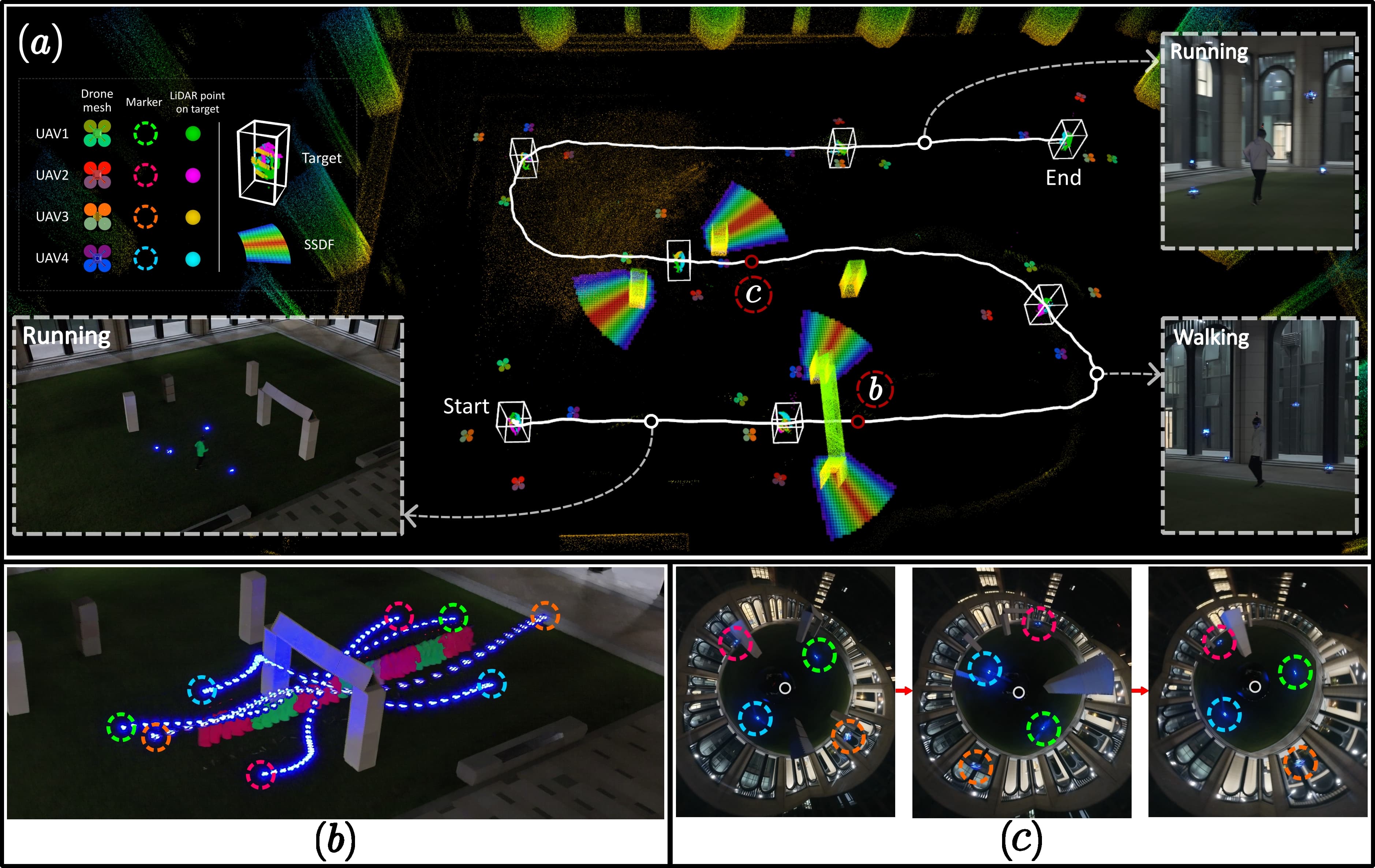}
  \caption{Four drones cooperatively track a human runner as the target. \textbf{(a)} The white curve depicts the runner's route. Three keyframes along the path are highlighted by the adjacent image snapshots, and two spots, labeled $b$ and $c$, are marked for detailed illustration in subfigures (b) and (c). \textbf{(b)} A composite image captures the swarm flexibly compressing its tracking distribution as the target runs through a gate at spot $b$. \textbf{(c)} A series of 360° camera snapshots shows the swarm rotating the distribution to prevent occlusion as the target walks through the pillar obstacles at spot $c$.}
  \label{fig:FM_exp}
  \vspace{-0.3cm}
\end{figure*}

To further validate our method's practicality, we deploy the swarm system 
to track a human runner. The runner transitions between walking 
($\sim1~m/s$) and running ($\sim2.5~m/s$). The swarm uses the same LiDAR 
setup as Sec.~\ref{sec:Swarm Tracking in Dense Forest}. The runner wears 
a high-reflectivity vest for fast LiDAR detection. Fig.~\ref{fig:FM_exp}(a) 
shows the scenario and target route. Fig.~\ref{fig:FM_exp}(b) is a composite 
image recording the swarm motion when the target runs through a gate. To 
preserve target visibility, the trackers flexibly compress the swarm 
distribution to traverse the constrained gateway and then elastically 
resume the tetrahedron tracking formation in open space, demonstrating 
swarm adaptability. Fig.~\ref{fig:FM_exp}(c) presents Insta360 snapshots when 
the target walks through pillar obstacles. Driven by SSDFs, the swarm 
rotates the distribution to avoid occlusions. Full visibility throughout 
walking and running phases confirms the system's reactive adaptation to 
target velocity changes.

\subsection{Swarm Tracking with Dynamic Joining and Leaving}
\label{Swarm Tracking with Dynamic Joining and Leaving}

\begin{figure*}[!t]
  \centering
  \includegraphics[width=0.99\textwidth]{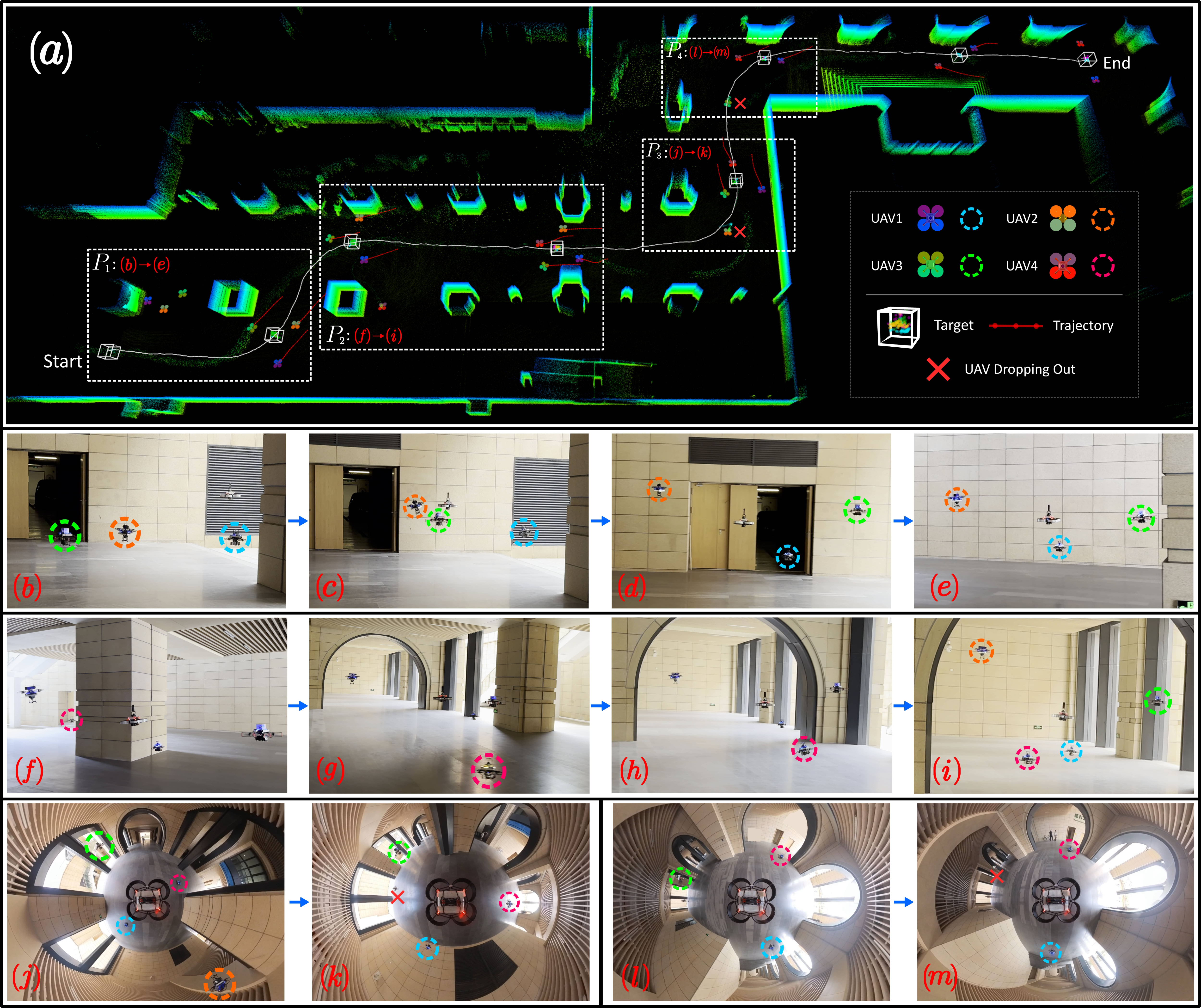}
  \caption{Swarm tracking experiment with dynamic joining and leaving. \textbf{(a)} The white curve shows the target drone's route. Four areas, labeled as $P_1$ to $P_4$, are marked for further illustration in the following subfigures. \textbf{(b)-(e)}: As the target enters area $P_1$, the swarm of UAV1-3 detects the target, starts cooperative tracking, and forms a regular triangle distribution. \textbf{(f)-(i)}: In $P_2$, UAV4 detects the target and initially performs solo tracking in (f)-(g). After the online calibration is complete in (h), UAV4 joins the swarm and then forms a tetrahedron encirclement with its teammates. \textbf{(j)-(k)}: In $P_3$, the swarm reconfigures the distribution from a tetrahedron to a regular triangle after UAV2 drops out. \textbf{(j)-(k)}: In $P_4$, the swarm transitions to collinear encirclement after UAV3 leaves.}
  \label{fig:FT_exp}
  \vspace{-0.3cm}
\end{figure*}

To demonstrate dynamic swarm reconfigurability, we conduct an 
experiment where teammates join and leave during live target tracking. 
A quadrotor marked by high-reflectivity tapes serves as the target, 
pursued by four drones (UAV1-4) with heterogeneous LiDARs: UAV1 and 
UAV4 (upward Mid360), UAV2 (downward Mid360), and UAV3 (Avia). 
Fig.~\ref{fig:FT_exp} records the tracking mission.

UAV1-3 complete swarm initialization~\cite{zhu2024swarm} in area $P_1$ 
to form a three-drone swarm, while UAV4 starts as an isolated agent 
in area $P_2$. When the target enters $P_1$, UAV1 first detects it 
and immediately shares its measurements with UAV2 and UAV3 via UDP, 
enabling coordinated tracking despite their FOV limitations 
(Fig.~\ref{fig:FT_exp}(b)-(c)). Driven by the distribution cost, 
the three drones form a triangular encirclement 
(Fig.~\ref{fig:FT_exp}(d)-(e)). As the target enters $P_2$, UAV4 detects it and begins solo tracking, 
treating other drones as dynamic obstacles since its extrinsic 
transformations are not yet calibrated with the swarm 
(Fig.~\ref{fig:FT_exp}(g)). Once the Swarm-LIO module completes online initialization, UAV4 joins the swarm and the four drones form a tetrahedron distribution (Fig.~\ref{fig:FT_exp}(h)-(i)). In area $P_3$, UAV2 is deliberately terminated to emulate agent 
failure. The system adaptively reconfigures into a triangle formation 
(Fig.~\ref{fig:FT_exp}(j)-(k)). In area $P_4$, UAV3 is further 
dropped, and the remaining two trackers form a collinear encirclement 
(Fig.~\ref{fig:FT_exp}(l)-(m)). In the experiment, our swarm maintains uninterrupted target tracking during dynamic membership changes, demonstrating the 
system's decentralized swarm reconfigurability. The inherent scalability of the proposed swarm tracking costs enables automatic adaptation to swarm-size variations without any hardcoded rules. The dropouts further verify the system's robustness and fault tolerance. 

\subsection{Communication Bandwidth}
Table~\ref{table:bandwidth} summarizes the average per-drone 
bandwidth of three data types across all real-world experiments. Our system adopts a fully-connected unicast topology as 
required by the data sharing framework, where each drone 
transmits data individually to every teammate.

\begin{table}[h]
\centering
    \renewcommand\arraystretch{1.3}
    \caption{Average Per-Drone Communication Bandwidth}
    \label{table:bandwidth}
    \scalebox{0.96}{
    \begin{tabular}{c c c c c c c}
    \toprule
      \multirow{2}*{Data Type} & Exp & IX-B & IX-C & \multicolumn{3}{c}{IX-D} \\
      \cmidrule(lr){5-7}
        & Size $N$ & 4 & 4 & 2 & 3 & 4 \\
       \hline
      \multirow{2}*{\makecell[c]{Target \\ Measurement}} 
        & TX(KB/s) & 1.23 & 1.37 & 0.45 & 0.92 & 1.39 \\
        & RX(KB/s) & 0.99 & 1.12 & 0.41 & 0.75 & 1.10 \\
       \hline
      \multirow{2}*{\makecell[c]{Planned \\ Trajectory}} 
        & TX(KB/s) & 3.40 & 3.37 & 1.15 & 2.26 & 3.39 \\
        & RX(KB/s) & 3.09 & 2.96 & 0.96 & 1.92 & 3.02 \\
       \hline
      \multirow{2}*{\makecell[c]{Local Map \\ Data}} 
        & TX(KB/s) & 67.90 & 39.76 & 21.57 & 43.37 & 59.95 \\
        & RX(KB/s) & 61.16 & 35.83 & 19.34 & 39.86 & 53.53 \\
       \hline
    \end{tabular}
    }
\end{table}

\section{Conclusion and Discussion}
\label{sec:conclusion}

This paper presented a visibility-aware cooperative tracking system 
for decentralized LiDAR-based swarms, featuring SSDF-based occlusion 
representation, differentiable FOV and swarm distribution metrics, and a two-stage planning framework. Real-world and simulation experiments validated robust 
swarm tracking performance in complex environments.

Several limitations remain. Visibility and formation 
constraints are enforced as soft objectives and may be 
temporarily violated in dense scenes. The constant-velocity 
target prediction can become inaccurate under aggressive 
maneuvers, and the unimodal Kalman filter cannot represent 
multiple hypotheses during prolonged occlusion. Communication 
losses may affect swarm coordination. In addition, while the 
planning framework is sensor-agnostic, the current LiDAR-based 
implementation relies on high-reflectivity markers for target 
detection. Future work will explore learning-based target 
motion prediction for aggressive maneuvers, formal verification 
for stronger constraint guarantees, robust communication 
protocols for larger swarm scales, non-additive 
visibility-weighted swarm distribution formulations that unify 
occlusion avoidance and spatial coordination, uncertainty-aware 
planning that incorporates target estimation covariance, and 
adaptation to vision-based sensing with pyramidal camera 
frustums.

\bibliography{references}

\end{document}